\pdfoutput=1
\documentclass[11pt]{article}

\usepackage{acl}

\usepackage{times}
\usepackage{latexsym}
\usepackage{graphicx}
\usepackage{graphics}
\usepackage{caption}
\usepackage{subcaption}

\usepackage{amssymb}
\usepackage{bbm}
\usepackage[T1]{fontenc}
\usepackage[utf8]{inputenc}

\usepackage{microtype}
\usepackage{amsmath,amsthm,amsfonts,amssymb,bm,stmaryrd,bbm}
\DeclareMathOperator*{\argmax}{arg\,max}

\title{On Isotropy, Contextualization and Learning Dynamics of Contrastive-based Sentence Representation Learning}

\author{Chenghao Xiao \quad Yang Long \quad Noura Al Moubayed\\
        Department of Computer Science\\ Durham University\\\texttt{\{chenghao.xiao,yang.long,noura.al-moubayed\}@durham.ac.uk}}
\begin{document}
\maketitle
\begin{abstract}

Incorporating contrastive learning objectives in sentence representation learning (SRL) has yielded significant improvements on many sentence-level NLP tasks. However, it is not well understood why contrastive learning works for learning sentence-level semantics. In this paper, we aim to help guide future designs of sentence representation learning methods by taking a closer look at contrastive SRL through the lens of isotropy, contextualization and learning dynamics. We interpret its successes through the geometry of the representation shifts and show that contrastive learning brings isotropy, and drives high intra-sentence similarity: when in the same sentence, tokens converge to similar positions in the semantic space. We also find that what we formalize as "spurious contextualization" is mitigated for semantically meaningful tokens, while augmented for functional ones. We find that the embedding space is directed towards the origin during training, with more areas now better defined. We ablate these findings by observing the learning dynamics with different training temperatures, batch sizes and pooling methods.

\end{abstract}

\section{Introduction}

Since vanilla pre-trained language models do not perform well on sentence-level semantic tasks, Sentence Representation Learning (SRL) aims to fine-tune pre-trained models to capture semantic information \cite{reimers2019sentence,li2020sentence,gao2021simcse}. Recently, it has gradually become \textit{de facto} to incorporate contrastive learning objectives in sentence representation learning \cite{yan2021consert,giorgi2021declutr,gao2021simcse,wu2022infocse}.

Representations of pre-trained contextualized language models \cite{peters-etal-2018-deep,devlin2019bert,liu2019roberta} have long been identified not to be isotropic, i.e., they are not uniformly distributed in all directions but instead occupying a narrow cone in the semantic space \citep{ethayarajh2019contextual}. This property is also referred to as the representation degeneration problem \cite{gao2018representation}, limiting the expressiveness of the learned models. The quantification of this characteristic is formalized, and approaches to mitigate this phenomenon are studied in previous research \citep{mu2018all,gao2018representation,cai2020isotropy}.

\begin{figure}
\begin{center}
\includegraphics[width=7.7cm]{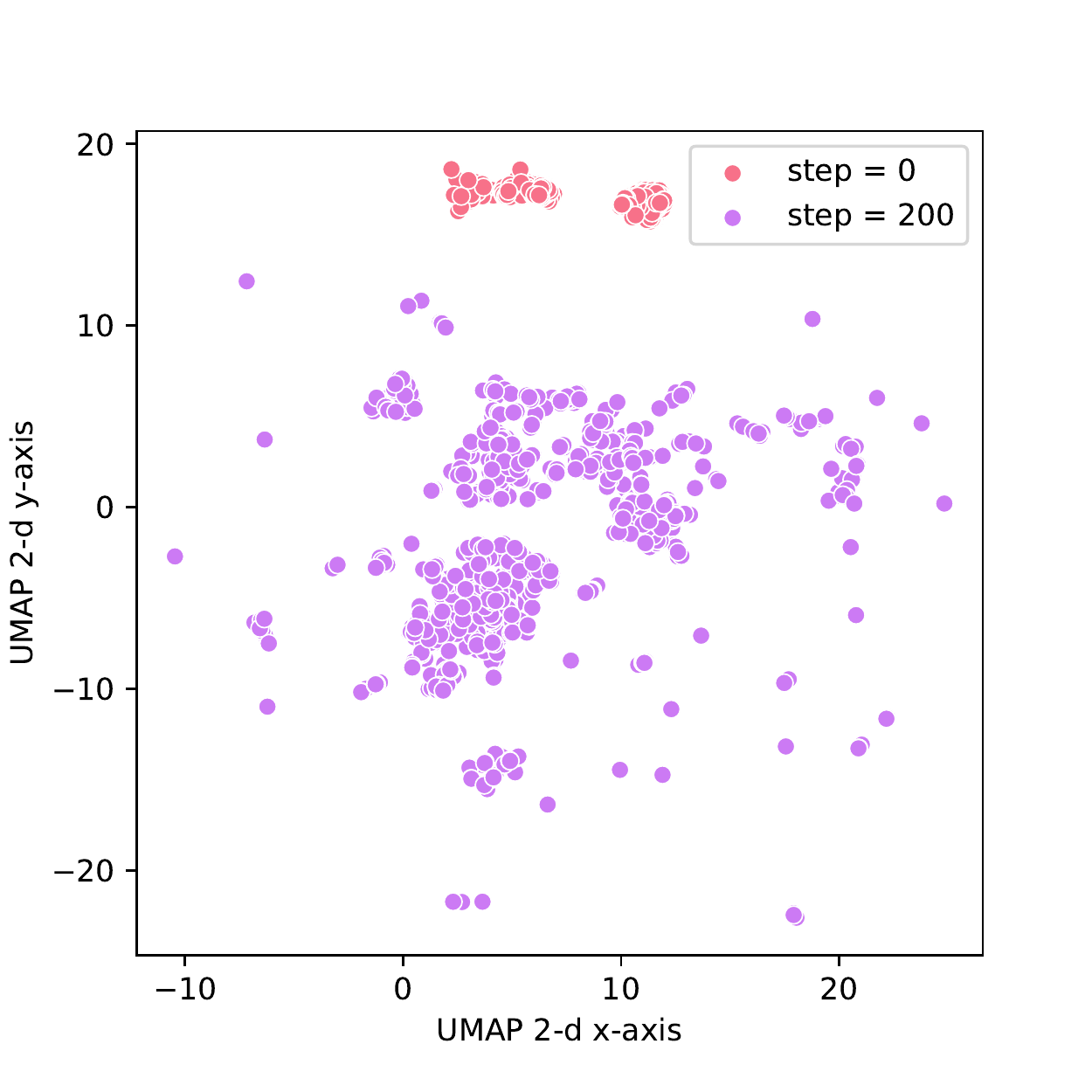}
\caption{Expanded semantic space produced by contrastive learning (CL), visualized with UMAP. At the beginning of training, all embeddings occupied a narrow cone. After 200 steps of fine-tuning with a contrastive loss, they spread out to define a larger semantic space.} \label{expanded semantic space}
\end{center}
\end{figure}

The concept of learning dynamics focuses on what happens during the continuous progression of fine-tuning pre-trained language models. This has drawn attention in the field \cite{merchant2020happens,hao2020investigating}, with some showing that fine-tuning mitigates the anisotropy of embeddings \cite{rajaee2021does}, to different extent according to the downstream tasks. However, it is argued that the performance gained in fine-tuning is not due to its enhancement of isotropy in the embedding space \cite{rajaee2021does}. Moreover, little research is conducted on isotropy of sentence embedding models, especially contrastive learning-based sentence representations.

Vanilla Transformer models are known to underperform on sentence-level semantic tasks even compared to static embedding models like Glove \cite{pennington2014glove,reimers2019sentence}, whether using the [cls] token or averaging word embeddings in the output layer. Since \citet{reimers2019sentence} proposed SBERT, it has become the most popular Transformers-based framework in sentence representation tasks. The state-of-the-art is further improved by integrating contrastive learning objectives \citep{yan2021consert,gao2021simcse,wu2022infocse}. The other line of works concern post-processing of embeddings in vanilla language models \citep{li2020sentence,su2021whitening,huang2021whiteningbert} to attain better sentence representations.

% Transformers-based language models often underperform on sentence modeling even compared to static embedding models \cite{pennington2014glove,reimers2019sentence}. Since \citet{reimers2019sentence} proposed SBERT, SOTA is further improved by integrating CL objectives \citep{yan2021consert,gao2021simcse,wu2022infocse}. Other works attain better sentence representations as post-processing \citep{li2020sentence,su2021whitening,huang2021whiteningbert}. 

Learning dynamics in fine-tuning was previously investigated, revealing isotropy shifts in the process \cite{rajaee2021does,gao2021simcse}, but few studies have systematically investigated relevant pattern shifts in sentence representation models, and none has drawn connections between these metrics and the performance gains on sentence-level semantic tasks. While some implicitly studied this problem by experimenting on NLI datasets \cite{rajaee2021does,merchant2020happens,hao2020investigating}, we argue that a more extensive study on the geometry change during fine-tuning SOTA sentence embedding models with contrastive objectives is neccessary.

In this work, we demystify the mechanism of why contrastive fine-tuning works for sentence representation learning.\footnote{\href{https://github.com/gowitheflow-1998/on-isotropy-of-contrastive-SRL/}{Our code is publicly available.}} Our main findings and contributions are as follows:

\begin{itemize}
    \item Through measuring isotropy and contextualization-related metrics, we uncover a previously unknown pattern: contrative learning leads to extremely high intra-sentence similarity. Tokens converge to similar positions when given the signal that they appear in the same sentence.
    \item We find that functional tokens fall back to be the "entourage" of semantic tokens, and follow wherever they travel in the semantic space. We argue that the misalignment of the "spurious contextualization change" between semantic and functional tokens may explain how CL helps capturing semantics. 
    \item We ablate all findings by analyzing learning dynamics through the lens of temperature, batch size, and pooling method, not only to validate that the findings are not artifacts to certain configurations, but also to interpret the best use of these hyperparamaters. 

\end{itemize} 

Our study offers fundamental insights into using contrastive objectives for sentence representation learning. With these, we aim to shed light on future designs of sentence representation learning methods.

\section{Isotropy and Contextualization Analysis of Contrastive-based Sentence Embedding models}
\label{section: isotropy analysis}
% In this section, we analyze isotropy-related metrics on sentence representation models that achieve SOTA performance on semantic textual similarity tasks, and their vanilla versions. We aim to decompose the changes of these models in the process of fine-tuning on sentence embedding tasks with CL objective, through comparing their metric values before and after fine-tuning.

\subsection{Preliminary}

Anisotropy of token embeddings produced by pre-trained language models has drawn attention in the field, and been validated both theoretically and empirically \cite{gao2018representation,ethayarajh2019contextual,cai2020isotropy,timkey2021all}.

For an anisotropic model, the embeddings it encodes have a high expected value of pair-wise cosine similarity: $\mathbb{E}_{u,v \in S}cos(u,v) >> 0$, where $u$ and $v$ are contextualized representations of tokens randomly sampled from corpus $S$. 

A contrastive learning objective to fine-tune a PLM on datasets that consist of sentence/document pairs is defined as follows:

\begin{equation}
\label{eqn:cse}
\ell_i = -\log \frac{e^{sim(e_i,e_i^+)/\tau}}{\sum_{j=1}^N e^{sim(e_i,e_j^+)/\tau}},
\end{equation}

where $e_i$ and $e_i^+$ denote embeddings of a sentence/document pair, whose cosine similarity is to be maximized, while all $e_j^+$ in a same training batch when $j \neq i$ is to be pushed further from $e_i$. 

The central question posed in this paper revolves around the mechanism involved in the contrastive learning process that diminishes anisotropy, leading to an isotropic model. If anisotropy is neutralized, we would observe a new mathematical expectation of cosine similarity, represented by $\mathbb{E}_{u,v \in S}cos(u,v) \approx 0 $. However, the precise process and the underlying mechanism that facilitate this transition remain the key questions we aim to address.

% The central question here: what is the mechanism of the phenomenon of anisotropy getting mitigated/removed in the contrastive learning process, making a model isotropic? This is manifested by a new expected value of  $\mathbb{E}_{u,v \in S}cos(u,v) \approx 0 $.

Therefore, metrics such as self-similarity of same tokens in different contexts, and intra-sentence similarity of different tokens in the same context, are pertinent. More importantly, we could further trace the contextualization shift that brings mitigated anistropy to word type, i.e., are functional words and semantic words less/more contextualized after contrastive learning? We show that, this finding could potentially attribute to the performance gain on sentence-level semantic tasks brought by contrastive fine-tuning. 

\subsection{Metrics}

We adopt the metrics defined in \citet{ethayarajh2019contextual}, who studied the extent to which word representations in pre-trained ELMo, BERT, and GPT-2 are contextualized, taking into consideration their anisotropy baselines. We reimplement the computation on self-similarity, intra-sentence similarity, and anisotropy baselines. We then break the similarity measures down into dimension level to inspect whether certain rogue dimensions \cite{timkey2021all} dominate these metrics and therefore making the similarity measures only artifacts of a small set of dimensions.

\paragraph{Self Similarity:} Self similarity measures the similarity among different contextualized representations of a token across different contexts. Higher self-similarity indicates less contextualization. Given a token $x$, we denote the set of token embeddings of $x$ contextualized by different contexts in corpus $S$ as $S_{\vec{\mathcal{X}}}$. Self similarity is then defined as the empirical mean of pair-wise cosine similarity of contextualized embeddings of token $x$ in all these contexts:

\begin{equation}
\begin{aligned}
\label{eqn:expected ith cos}
selfsim(x) 
% & = \mathbb{E}[\bar{ss}] \\
%   & = \mathbb{E}_{u,v \in S}[cos(u_i,v_i)] \\
& \triangleq  \mathbb{E}_{u,v \in S_{\vec{\mathcal{X}}}} [\bar{cos}(u,v)]
% & \approx \frac{1}{k} \sum_{i=1}^{k} \bar{cos}(u_i, v_i)
\end{aligned}
\end{equation}

\paragraph{Intra-sentence Similarity:} By contrast, intra-sentence similarity measures the similarity across tokens in the same context. 

Given a sentence $s$ with $n$ tokens $x_{i 
\in \{1, 2, ..., n\}}$, we first attain sentence representation $\vec{s}$ by mean-pooling, i.e., averaging all token embeddings $\vec{x_i}$. Intra-sentence similarity is then defined as the average cosine similarity between token representations $\vec{x_i}$ and the sentence representation $\vec{s}$.

\begin{equation}
\begin{aligned}
\label{eqn:expected ith cos}
\vec{s} 
& \triangleq  \frac{1}{n}\sum_{x_i \in s} \vec{x_i}\\
intrasim(s) 
& \triangleq  \frac{1}{n}\sum_n{cos}(\vec{x_i},\vec{s})
\end{aligned}
\end{equation}

Intra-sentence similarity provides a quantitative measure of the extent to which tokens in the same sentence are similar, allowing us later to derive insights on: whether token representations would converge in the semantic space only because they appear in a same sentence.

\paragraph{Anisotropy Baselines:}  While self and intra-sentence similarity are computed given the restrictions of respectively 1) same word in different contexts 2) different words in the same context, these values are not reflective of the general distribution across different words and different contexts. 

In line with \citet{ethayarajh2019contextual}, we adjust the above two metrics by substracting the anisotropy baseline of a model from them, i.e., average cosine similarity between randomly sampled tokens from different contexts as defined in preliminary. 
% We therefore subtract from each similarity measure its respective anisotropy baseline.

%For instance, if the average cosine similarity of two random word representations is 0 given enough sampling from the corpus (meaning the word vectors are fully isotropic), then a word is not highly contextualized if its self similarity is say 0.7. By contrast, if two random words have an average cosine similarity of 0.7, then a self similarity of 0.7 means that a word is decently contextualized - It is not even more similar to itself in other contexts than to a random word in a random context.

\paragraph{Dimension-level Inspection of the Metrics} Due to the fact that cosine similarity is highly sensitive to outlier dimensions, we inspect whether the outcomes of the above measurements are only artifacts of these dimensions, i.e. rogue dimensions \cite{timkey2021all}.

Formally, the cosine similarity of two embeddings is defined as: $cos(u,v) = \frac{u \cdot v}{\left\| u \right\|\left\| v \right\|},$ where $u$ and $v$ are two embeddings to measure against. Since the term $u \cdot v$ is just a sum of the element-wise dot product of the $i^{th}$ dimension of the embeddings, it is convenient to inspect the contribution each dimension makes to the global similarity: $cos(u,v) = \sum_{i=1}^d \frac{u_i v_i}{\left\| u \right\|\left\| v \right\|}$.

Given a set $S$ that consists of $n$ randomly sampled representations, the expected contribution of the $i^{th}$ dimension in a model to a similarity metric could be approximated as:

\begin{equation}
\begin{aligned}
\label{eqn:expected ith cos}
cos_i 
%   & = \mathbb{E}_{u,v \in S}[cos(u_i,v_i)] \\
  & = \mathbb{E}_{u,v \in S} \frac{u_i v_i}{\left\| u \right\|\left\| v \right\|},
\end{aligned}
\end{equation}

By breaking the global metrics down to dimension level, whether the output of a metric is a global property of all embeddings in the language model or is only dominated by a set of rogue dimensions $D$ could be inspected by whether $\sum_{i \in D} cos_i >> \frac{\|D\|}{d}\mathbb{E}_{u,v \in S} cos(u,v)$, with $d$ being the dimensionality of word embeddings.

Nonetheless, we could mathematically derive that, dominating dimensions dominate corpus-level similarity metric computations mostly because of their high average distances to the origin at the corresponding dimensions. However, if the values in these dimensions do not have high variation, then eliminating the top $\| D \|$ of these dimensions from the embeddings would not significantly bring semantic shifts to the original representations and therefore would not affect the corresponding relative similarity relationship between sentence pairs. %For instance, if the 768-d embeddings of the phrases "good sentence embedding" and "legit sentence representation" yield highest pair-wise cosine similarity throughout the corpus, removing the 5 "rogue scalers" from both (making each phrase a 763-d embedding) would still provide them highest similarity compared to other phrases in the corpus, in this case. By contrast, if the variations of these rogue dimensions are also among the highest, the above example would not likely to hold.

Therefore, we will also need to inspect whether there is a misalignment between the existence of the rogue dimensions, and their actual impact on informativity \cite{timkey2021all}. Given a $f(t, k)$ that maps a token $t$ to its representation, with top $k$ rogue dimensions eliminated, we could compare the correlation between similarity measures yielded by the original representatations and those with top-k rogue dimensions removed. Formally, given:
\begin{equation}
\begin{aligned}
\label{eqn: original}
cos_{original}(\mathcal{O}) = cos_{x,y \in \mathcal{O}}(f(x, 0),f(y,0))
\end{aligned}
\end{equation}
\begin{equation}
\begin{aligned}
\label{eqn: post}
cos_{post}(\mathcal{O}) = cos_{x,y \in \mathcal{O}}(f(x, k),f(y,k)),
\end{aligned}
\end{equation}
we compute: $r = Corr[cos_{original}, cos_{post}]$, which is an indicator of the "authenticity" of the representations left without these rogue dimensions.

With the corresponding dimension-level inspections of the three metrics, we could take a step further to investigate whether fine-tuning a vanilla language model to sentence embedding tasks with the contrastive objective mitigates the dominance of rogue dimensions.

\subsection{Models}

We analyze two models that achieve SOTA performances on sentence embedding tasks and semantic search tasks, \textit{all-mpnet-base-v2}
\footnote{https://huggingface.co/sentence-transformers/all-mpnet-base-v2}
and \textit{all-MiniLM-L6-v2}.
\footnote{https://huggingface.co/sentence-transformers/all-MiniLM-L6-v2}
They have both been fine-tuned with a contrastive loss on 1B+ document pairs, with the goal of predicting the right match to a document $d_i$ given its ground-true match $d_i^+$ and the rest of the in-batch $d_j^+$ as natural negative examples. The prediction is conducted again reversely with $d_i^+$, $d_i$ and other in-batch $d_j$. The loss is averaged for these two components for every batch. The representation of each document $d$ is by default the mean-pooled embedding of each token.

We compare the results to their vanilla versions, \textit{mpnet-base} \citep{song2020mpnet} and \textit{MiniLM}\footnote{
% Notably, the original implementation of MiniLM has 12 layers, while we use a 6-layer version (by taking every second layer) to align the analysis with its fine-tuned counterpart. The model could be found in 
https://huggingface.co/nreimers/MiniLM-L6-H384-uncased.  Notably, we use a 6-layer version.} \citep{wang2020minilm} to get a closer look to the initial state of their corresponding pre-trained counterparts, and how the metrics change after fine-tuning on the goal of getting better sentence and document-level representations.

\subsection{Data}

We use STS-B \cite{cer-etal-2017-semeval}, which comprises a selection of datasets from the original SemEval datasets between 2012 and 2017. We attain the dataset through Hugging Face Datasets\footnote{https://huggingface.co/datasets/stsb\_multi\_mt}. Notably, the models that we are looking at were not exposed to these datasets during their training. Therefore, the pattern to be found is not reflective of any overfitting bias to their training process. 

We use the test set and only use sentence 1 of each sentence pair to prevent the potential doubling effect on self-similarity measure, i.e., providing tokens with one more sentence where they are in the similar contexts. 
% Such similar contexts would offset the self-contextuality metric, which is supposed to be reflective of the distribution of their real-world usage. 
Following the description, 1359 sentences are selected as inputs. 
% to experiment on the contextuality and isotropy-related metrics, and the dimension-wise dominating pattern of the corresponding models.

\subsection{Result}

\label{ots results}

\begin{figure}[h]
\includegraphics[width=7.7cm]{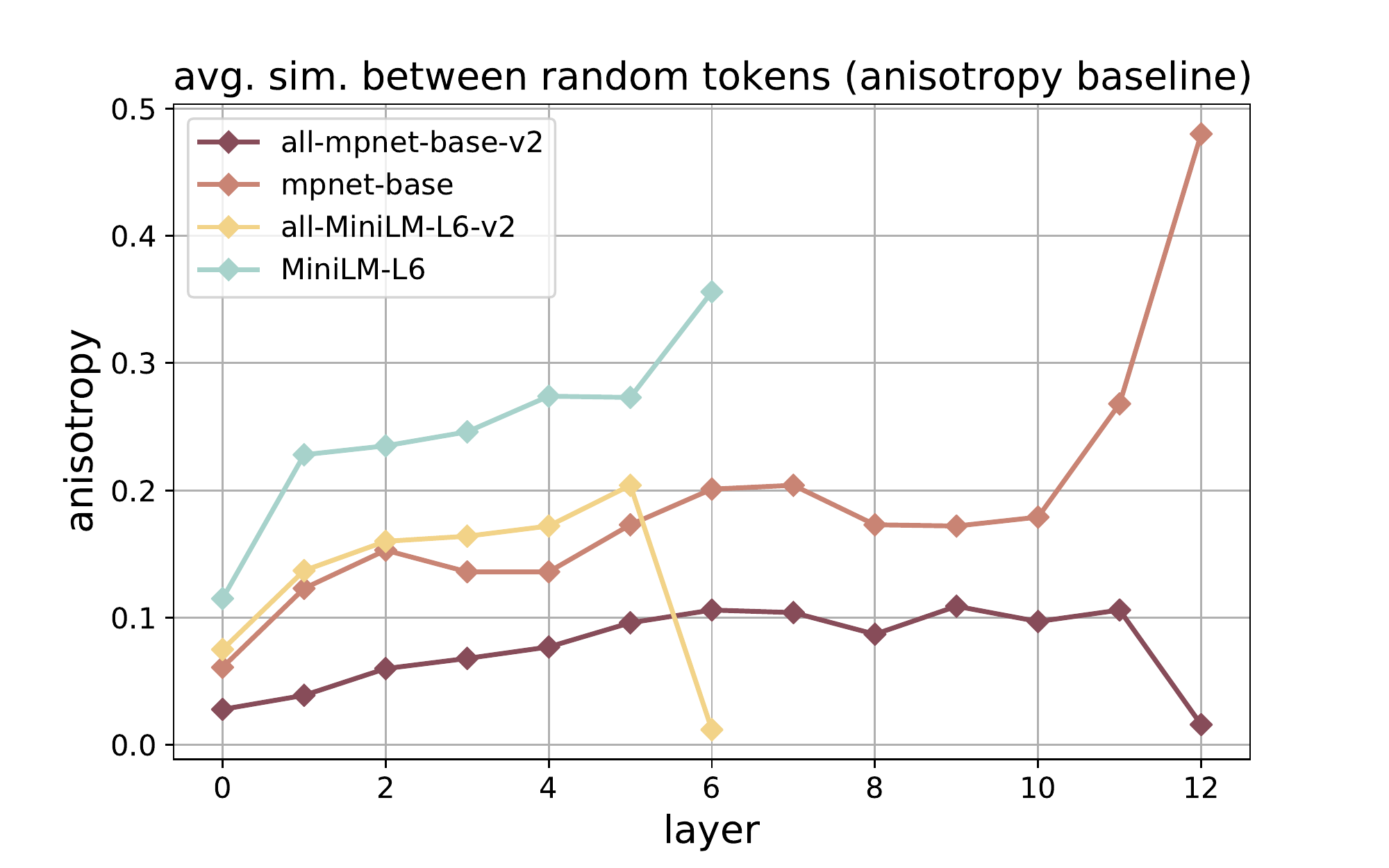}
\caption{Anisotropy baseline of models} \label{ots-ani-baseline}
\end{figure}

We show that after fine-tuning with contrastive loss, the anisotropy is almost eliminated in the output layer of both models, and is mitigated in the middle layers to different levels. This empirically validates the theoretical promise of uniformity brought by contrastive learning \cite{wang2020understanding,gao2021simcse} in the context of sentence representation learning (Figure~\ref{ots-ani-baseline}).

Complementing the enhanced isotropy, the average L2 norm of the randomly sampled token representations is also measured, showing a similar drastic shift in mostly the output layer of both models. Geometrically, the embeddings of tokens are pushed toward the origin in the output layer of a model, compressing the dense regions in the semantic space toward the origin, making the embedding space more defined with concrete examples of words (see also Figure~\ref{expanded semantic space}), instead of leaving many poorly-defined areas \cite{li2020sentence}. This property potentially contributes to models' performance gains on sentence embedding tasks.

\begin{figure}[h]
\includegraphics[width=7.7cm]{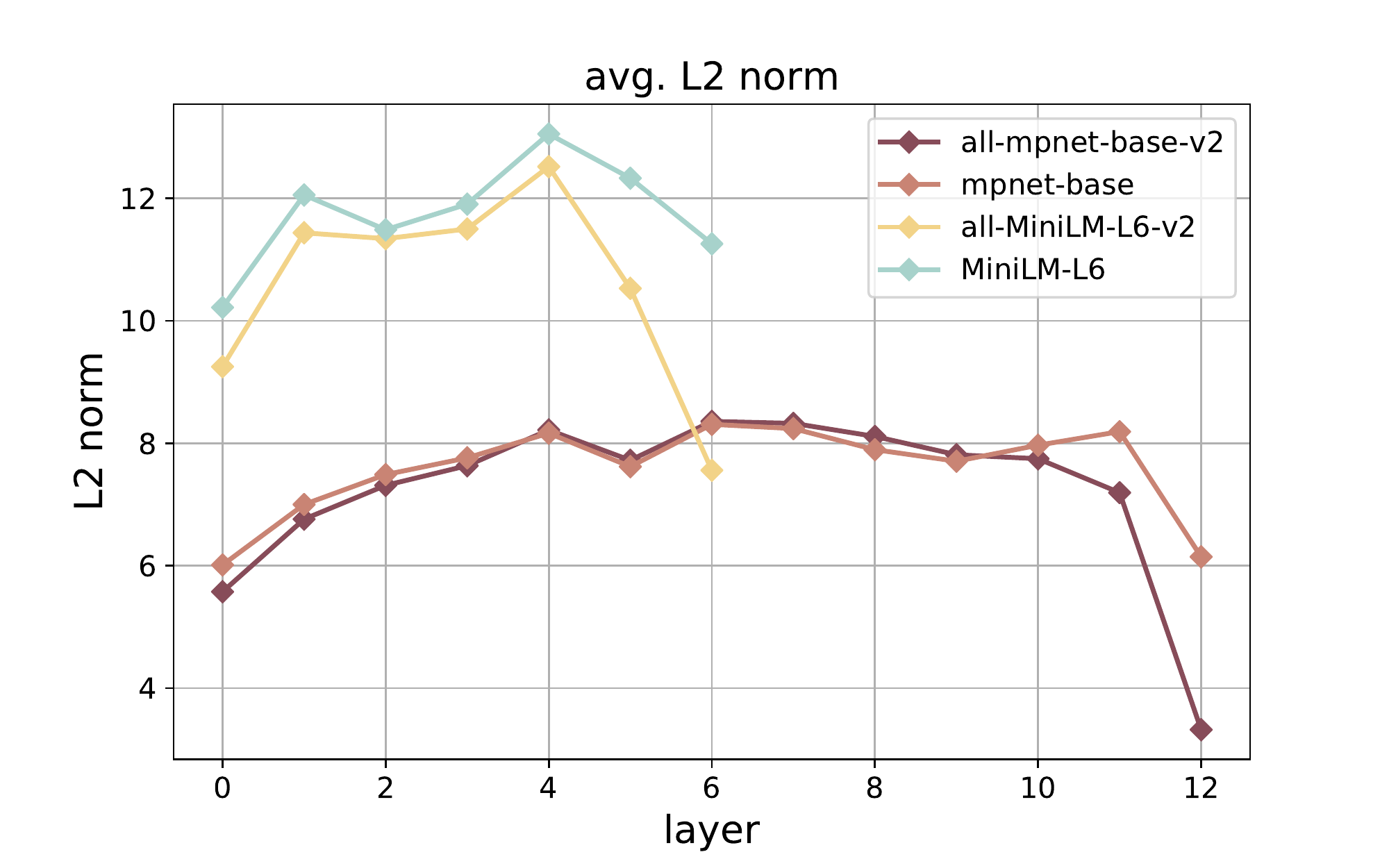}
\caption{Avg. L2 norm of embeddings} \label{ots-norm}
\end{figure}

\begin{figure}[h]
\includegraphics[width=7.7cm]{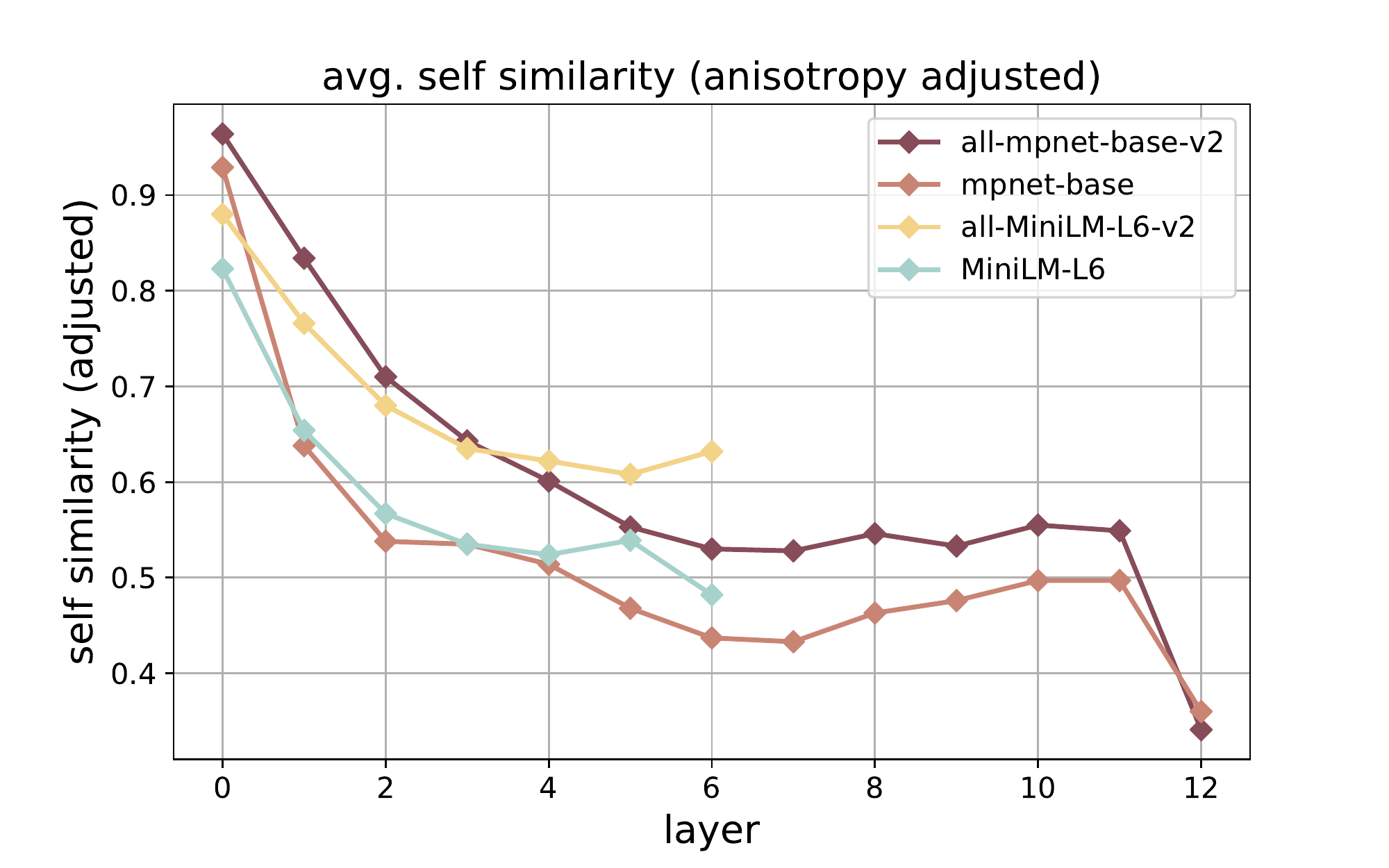}
\caption{Adjusted self similarity of tokens: each self similarity is adjusted by the anisotropy of the corresponding model} \label{ots-self-adjusted}
\end{figure}

\begin{figure}
\includegraphics[width=7.7cm]{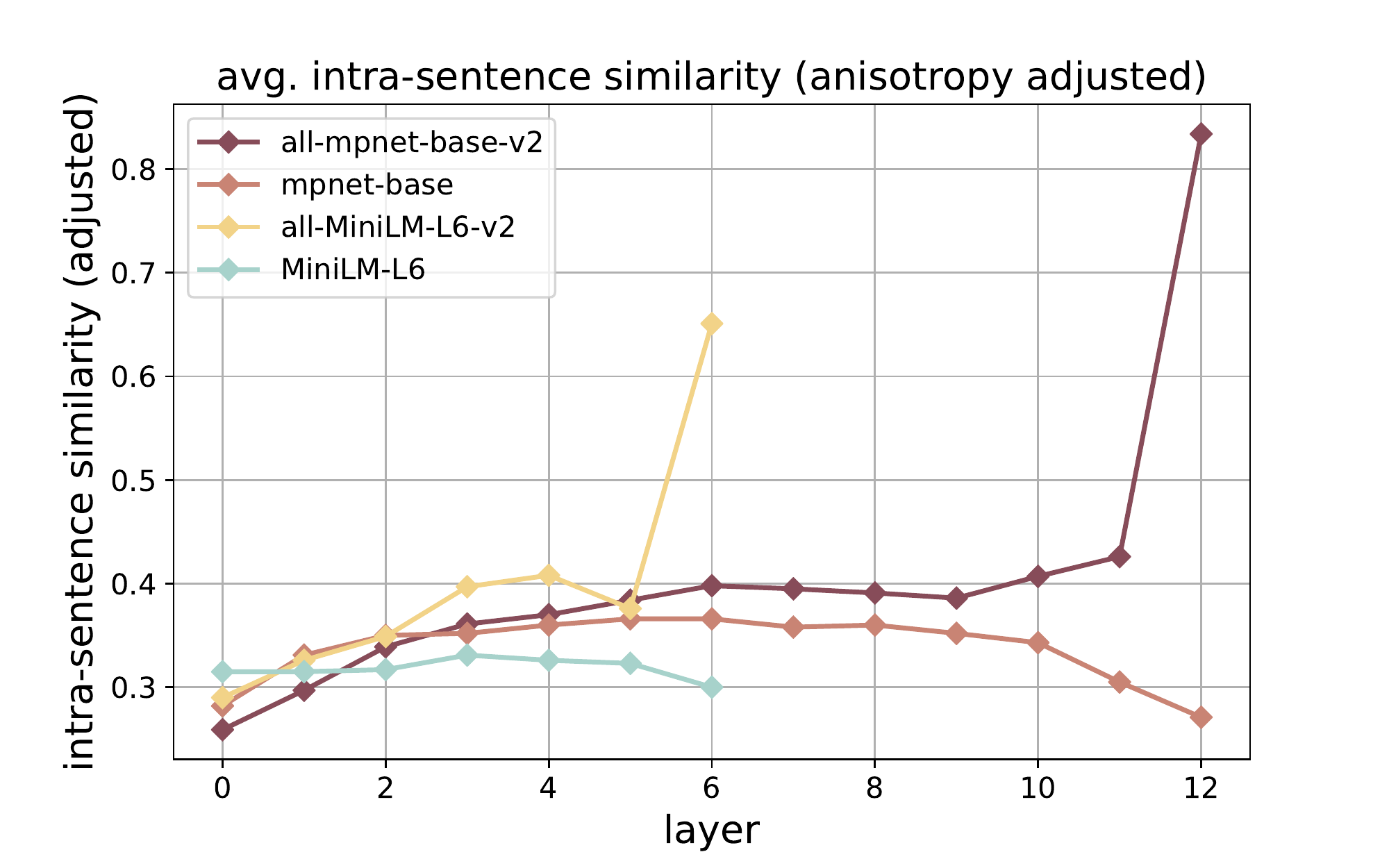}
\caption{Adjusted intra-sentence similarity of tokens: each intra-sentence similarity is adjusted by the anisotropy of the corresponding model} \label{ots-intra-adjusted}
\end{figure}

Figure~\ref{ots-self-adjusted} and Figure~\ref{ots-intra-adjusted} present respectively the self similarity and intra-sentence similarity of models adjusted (subtracted) by their anisotropy baselines (Unadjusted measures in Appendix~\ref{sec:unadjusted measure appendix}).

As for the adjusted self similarity, we can see that the fine-tuned models generally show higher self similarities across contexts (meaning tokens are less contextualized after fine-tuning) in all layers, except for the output layer of the fine-tuned mpnet. However, in general there does not exist a large difference on this metric (See why in Section~\ref{sec:frequency bias}).

We observe that intra-sentence similarity dramatically goes up in the output layer after contrastive fine-tuning. In the output layer of fine-tuned mpnet, the intra-sentence similarity reaches 0.834 (adjusted), meaning that tokens are 83.4\% similar to one another if they appear in a same sentence. Since this pattern does not exist in the vanilla pre-trained models, the pattern is a unique behavior that accompanies the performance gain brought by contrastive learning. We argue that given contrastive examples and the goal of distinguishing between similar and non-similar in each batch, the model learns to provide more intense cross-attention among elements inside an input, and thus could better assign each example (sentence/document) to a unique position in the semantic space. With mean-pooling and positive pairs, the model learns to decide important tokens in a document $d_i$, in order to align with its paired document $d_i^+$, and other secondary tokens are likely to \textbf{imitate} the embeddings of these important tokens because they need to provide an average embedding together to match with their counterpart (In Appendix~\ref{sec: pooling method} we conduct an ablation study with other pooling methods). Further, with limited space in the now compressed space, inputs have now learned to converge to one another to squeeze to a point while keeping its semantic relationship to other examples. Therefore, we reason that, the unique behavior of this "trained intra-sentence similarity" is highly relevant to the models' enhanced performance on sentence-level semantic tasks.

\begin{table}[h]
\centering
\begin{tabular}{llll}
\hline
\textbf{Model} & \textbf{Top 1} & \textbf{Top 2} & \textbf{Top 3}\\
\hline
\hline
mpnet\textsubscript{vanilla} & .548  & .723 & .741\\
mpnet\textsubscript{fine-tuned} & .005  & .010 & .014\\
\hline
minilm\textsubscript{vanilla} & .081 & .129 & .163\\
minilm\textsubscript{fine-tuned} & .008 & .014 & .020 \\
\hline
\hline
 & \textbf{10\%} & \textbf{20
\%} & \textbf{50\%}\\
\hline
mpnet\textsubscript{vanilla} & 1 & 1 & 1\\
mpnet\textsubscript{fine-tuned} & 28  & 64 & 209\\
\hline
minilm\textsubscript{vanilla} & 2 & 5 & 31\\
minilm\textsubscript{fine-tuned} & 19 & 40 & 121 \\
\hline
\end{tabular}
\caption{\label{Rogue Analysis}
Dimension-wise inspection on vanilla and contrastive learning-based fine-tuned sentence representation models (last/output layer only). The upper part of the table presents the contribution percentage of the top 1 to 3 dominating dimensions. The lower part provides the number of top dimensions needed to account for \{10,20,50\}\% of similarity metric computation.
}
\end{table}

Complementing the global properties found above, we present in Table~\ref{Rogue Analysis} the dimension-level inspection on the measures. The analysis is conducted on self similarity. In line with previous work \citep{timkey2021all}, there exists a significantly unequal contribution among dimensions. This inequality is most pronounced in the vanilla mpnet, with the top 1 dimension (out of the total 768) contributing to almost 55\% of the similarity computation. After contrastive fine-tuning, this phenomena is largely removed, with dominating dimensions greatly "flattened" \citep{gao2021simcse}. For the fine-tuned mpnet, it now requires 209 (out of 768, 27.2\%) dimensions to contribute to 50\% of the metric computation, and for fine-tuned minilm, this number is 121 (out of 384, 31.5\%).

In Appendix~\ref{informativity}, we present the informativity analysis by removing top-k dominating dimensions, we see a reallocation of information after contrastive fine-tuning and a misalignment between dominance toward similarity computation and informativity.

% \section{Connecting Isotropy, Self and Intra-sentence Similarity with Frequency Bias}
\section{Connecting to Frequency Bias}
\label{sec:frequency bias}
The imbalance of word frequency has long been identified to be relevant to the anisotropy of trained embeddings \cite{gao2018representation}. This has been also empirically observed in pre-trained Transformers like BERT \cite{li2020sentence}. \citet{li2020sentence} draw connection between frequency bias and the unideal performance of pre-trained language models on STS tasks, through deriving individual words as connections of contexts, concluding that rare words fail to play the role of connecting context embeddings. \citet{rajaee2021does} show that when fine-tuning pre-trained language models under the setting of Siamese architecture on STS-b datasets, the frequency bias is largely removed, with less significant frequency-based distribution of embeddings. However, it is also pointed out that these trained models are still highly anisotropic, which as we showed in Section~\ref{ots results}, does not hold in the context of contrastive training, which, with sufficient data, has theoretical promise toward uniformity \cite{wang2020understanding,gao2021simcse}.

Therefore, it is of interest to see the corresponding behaviors of frequency bias shifts in the context of contrastive learning, and more importantly, how this correlates with our surprising finding on intra-sentence similarity.

\subsection{How Self Similarities Change for Frequent Words?}

Since word frequency has produced many problematic biases for pre-trained Transformer models, we would like to know whether contrastive learning eases these patterns. Thus, how the self-similarity measurement manifests for frequent words after the models are fine-tuned with the contrastive objective? Are they more/less contextualized now?

\paragraph{Validity of Measuring Self-Similarity Change}

We first define Self-Similarity Change and prove that this measurement is not prone to stochasticity in the training process.

The top 400 frequent tokens are first extracted from the constructed STS-b subset. Then, we measure the avg. self-similarity before and after fine-tuning for each word, adjusted for their anisotropy baseline. Formally, we define Self-Similarity Change (SSC) of a token as:
\begin{equation}
\begin{aligned}
\label{eqn: post}
ssc = (ss_{f} - ani_{f}) - (ss_{v} - ani_{v}),
\end{aligned}
\end{equation}

where $ss_{f}$, $ss_{v}$, $ani_{f}$ and $ani_{v}$ stand for self-similarity and anisotropy baseline of fine-tuned and vanilla models respectively.

To validate that this measurement is not a product of stochasticity occurs in training but a common phenomenon that comes with contrastive learning, we compute the Self-Similarity Change for every token using both \textit{mpnet} (vanilla $\&$ fine-tuned) and \textit{MiniLM} (vanilla $\&$ fine-tuned). If the statistics produced by both models show high correlation, then there exists a pattern that would affect how self-similarity changes for different tokens during contrastive fine-tuning. Otherwise, the changes are a product of randomness.

We iterate $n = 1$ to $400$ to compute the Pearson correlation of SSCs of the top $n$ tokens produced by both \textit{mpnet} and \textit{MiniLM} and find the position where these statistics correlate the most, which is:
% \begin{equation}
% \begin{aligned}
% \label{eqn: post}
$\argmax\limits_n({corr(ssc_{mpnet}[:n], ssc_{MiniLM}[:n])})$.
% \end{aligned}
% \end{equation}
Throughout the iteration, the top $204$ frequent tokens give the highest Pearson correlation, which reaches a surprisingly high number of $0.857$, validating the universal pattern for similarity shifts of frequent words. After inspection, we find that these are tokens that appear more than $9$ times in the $1359$ sentences. Notably, even the full set of $400$ tokens gives a correlation of over 0.8, again proving the robustness of this pattern for frequent words (Refer to Appendix~\ref{sec: SSC} for the full statistics of the validation).

\subsection{Reaching to the connection}
Table~\ref{frequent words self-sim change} provides a glimpse of the top 10 tokens (among the top 400 frequent tokens) that are now most more contextualized (with top negative self-similarity changes) and most less contextualized (with top positive self-similarity changes).

\begin{table}[h]
\centering
\begin{tabular}{lllll}
\hline
 &  \multicolumn{2}{c}{mpnet} & \multicolumn{2}{c}{minilm}\\
\hline
{}   & SS ($\downarrow$) & SS ($\uparrow$) & SS ($\downarrow$)   & SS ($\uparrow$)\\
0   &  has & onion & [SEP] & hands\\
1   &  is & piano & . & fire\\
2   &  ,  &  unfortunately   & ;  & run\\
3   &  '  & cow   & ?  & house\\
4   &  are & chair   & )  & japan\\
5   &  that  &  potato  & the  & hat\\
6   &  been & read  & an & ukraine\\
7   &  while & dow   & -  & jumping\\
8   &  was  &  guitar  & / & coffee\\
9   &  with & drums  & a  & points\\
\hline
\end{tabular}

\caption{\label{frequent words self-sim change}Top Self-Similarity Changes}
\end{table}

After contrastive fine-tuning, tokens that contribute more to the semantics (tokens that have POS like nouns and adjectives) are now more reflective of their real-world limited connotations - tokens like "onion" and "piano" are not supposed to be that different in different contexts as they are in pre-trained models. We formalize this as \textbf{"Spurious Contextualization"}, and establish that \textbf{contrastive learning actually mitigates this phenomena for semantically meaningful tokens}. We speculate that these tokens are typically the ones that provide aligning signals in positive pairs and contrastive signals in negative pairs.

By contrast, however, the spurious contextualization of stopwords is even augmented after contrastive learning. "Has" is just supposed to be "has" - as our commonsense might argue - instead of having $n$ meanings in $n$ sentences. We speculate that, \textbf{stopwords fall back to be the "entourage" of a document after contrastive learning}, as they are likely the ones that do not reverse the semantics and thus do not provide contrastive signals in the training. Connecting this to our finding on high intra-sentence similarity, we observe that given a sentence/docuemnt-level input, certain semantic tokens drive the embeddings of all tokens to converge to a position, while functional tokens follow wherever they travel in the semantic space. 

\section{Ablation Analysis}
\label{section: tempearature}
In this section, we provide a derivation to interpret the role of temperature in CL, inspiring the searching method of its optimal range. We also show that contrastive frameworks are less sensitive to batch size at optimal temperature for SRL, unlike in visual representation learning.
\subsection{Rethinking Temperature}
Given a contrastive learning objective: $\ell_i = -\log \frac{e^{sim(e_i,e_i^+)/\tau}}{e^{sim(e_i,e_i^+)/\tau} + \sum_{j=1}^N \mathbbm{1}_{\{ j \neq i\}} e^{sim(e_i,e_j^+)/\tau}},$ we first look at its denominator, where the goal is to minimize the similarity between the anchor $e_i$ and negative pairs $e_j$ when $j \neq i$:
\begin{equation}
\label{eqn:numerator}
e^{sim(e_i,e_j^+)/\tau} \in (\frac{1}{e}^{\frac{1}{\tau}}, e^{\frac{1}{\tau}})
\end{equation}
Let $x$ be $e^{sim(e_i,e_i^+)}$ we get:
\begin{equation}
\label{eqn:e to x}
e^{sim(e_i,e_j^+)/\tau} = x^{1/\tau}, x \in (\frac{1}{e}, e)
\end{equation}
If $\tau << 1$, as long as $x < 1$, $x^{1/\tau}$ shrinks exponentially. While when $x > 1$, $x^{1/\tau}$ explodes exponentially. Therefore, $x = 1$, or $sim(e_i,e_j^+) = 0$ when $i \neq j$ is an important threshold when negative pairs are to decide whether or not to further push away, and this "thrust", is exactly what temperature provides: In-batch negatives are not motivated to be too dissimilar under a lower temperature, since once the similarity reaches below 0, the exponent ${1/\tau}$ is already doing the job of making them exponentially vanishing in the denominator.

We analyze the upper bound and lower bound of $sim(e_i,e_j^+)$ under 0, giving us $sim(e_i,e_j^+) = 0$ and $sim(e_i,e_j^+) = -1$ for every $sim(e_i,e_j^+)$ in batch when $i \neq j$. For both cases we pair them with $sim(e_i,e_i^+) \to 1^-$ since positive pairs are drawn closer regardless. Therefore,

\begin{equation}
\begin{aligned}
\label{eqn:upper bound}
\ell_{upper bound}(\tau) 
& = -\log \frac{e^{sim(e_i,e_i^+)/\tau}}{e^{sim(e_i,e_i^+)/\tau} + \sum\limits^{n-1}{e^{0/\tau}}} \\
& = -\log \frac{e^{sim(e_i,e_i^+)/\tau}}{e^{sim(e_i,e_i^+)/\tau} + (n-1)}
,
\end{aligned}
\end{equation}
while given $sim(e_i,e_i^+) \to 1^-$,
\begin{equation}
\begin{aligned}
\label{eqn:lower bound}
\ell_{lower bound}(\tau)
& = -\log \frac{e^{sim(e_i,e_i^+)/\tau}}{e^{sim(e_i,e_i^+)/\tau} + \sum\limits^{n-1}{e^{-1/\tau}}} \\
& = -\log \frac{e^{(sim(e_i,e_i^+)+1)/\tau}}{e^{(sim(e_i,e_i^+)+1)/\tau} + (n-1)} \\
& \approx -\log \frac{e^{2*sim(e_i,e_i^+)/\tau}}{e^{2*sim(e_i,e_i^+)/\tau} + (n-1)}
\end{aligned}
\end{equation}
Therefore, $\ell_{lower bound}(2\tau) \approx \ell_{upper bound}(\tau)$.

We find that temperature affects making embeddings isotropic: to push in-batch negatives to the lower bound, the temperature needs to be twice as large than to push them to the upper bound. For example, if when temperature $= 0.05$, two sentences are pushed in training to have $-1$ cosine similarity, now given temperature  $= 0.025$, the gradient is only around enough to push them to have $0$ cosine similarity with each other. 

The findings suggest that searching for the optimal value of this hyperparameter using a base of 10, as empirically shown in previous research \cite{gao2021simcse}, may not be the most efficient approach. Instead, we argue that a base of 2 would be more appropriate, and even to conduct finer-grained searching when a range of upper bound temperature that is twice the lower bound temperature is found to provide adequate performance. 

Our analysis serves as a complementation to \citet{wang2021understanding}, who show that a lower temperature tends to punish hard-negative examples more (especially at the similarity range of $(0.5,1)$), while a higher temperature tends to give all negative examples gradients to a same magnitude. This provides more theoretical justification to our approximation, since at the similarity range of $(-1,0)$, all negative examples have gradients to the same magnitude \cite{wang2021understanding} regardless. We suggest that this range plays a main role in making the entire semantic space isotropic.

\subsection{Experiment Setup}

We use a vanilla mpnet-base \cite{song2020mpnet} as the base model, and train it on a concatenation of SNLI \cite{bowman2015large} and MNLI datasets \cite{williams2018broad}. In accordance with our analysis, for the temperature $\tau$ subspace we deviate from the commonly adopted exponential selection with a base of 10 (e.g., \citet{gao2021simcse}), but we analyze around the best value found empirically, with a base of 2, i.e., $\{0.025, 0.05, 0.1\}$. We provide the same analysis on $\{0.001, 0.01, 0.05, 0.1, 1\}$ in Appendix~\ref{sec:temperature search appendix} for comparison. To better illustrate the effect of temperature, we only use entailment pairs as positive examples, under supervised training setting. We do not consider using contradiction as hard negatives to distract our analysis, nor unsupervised settings using data augmentation methods such as standard dropout. We use all instances of entailment pairs as training set, yielding a training set of $314$k. We truncate all inputs with a maximum sequence length of 64 tokens. All models are trained using a single NVIDIA A100 GPU. We train the models with different temperatures for a single epoch with a batch size of $64$, yielding 4912 steps each, with $10\%$ as warm-up. We save the models every 200 steps and use them to encode the subset of STS-B we have constructed.

\subsection{Results}
\begin{figure}[h]
\includegraphics[width=7.7cm]{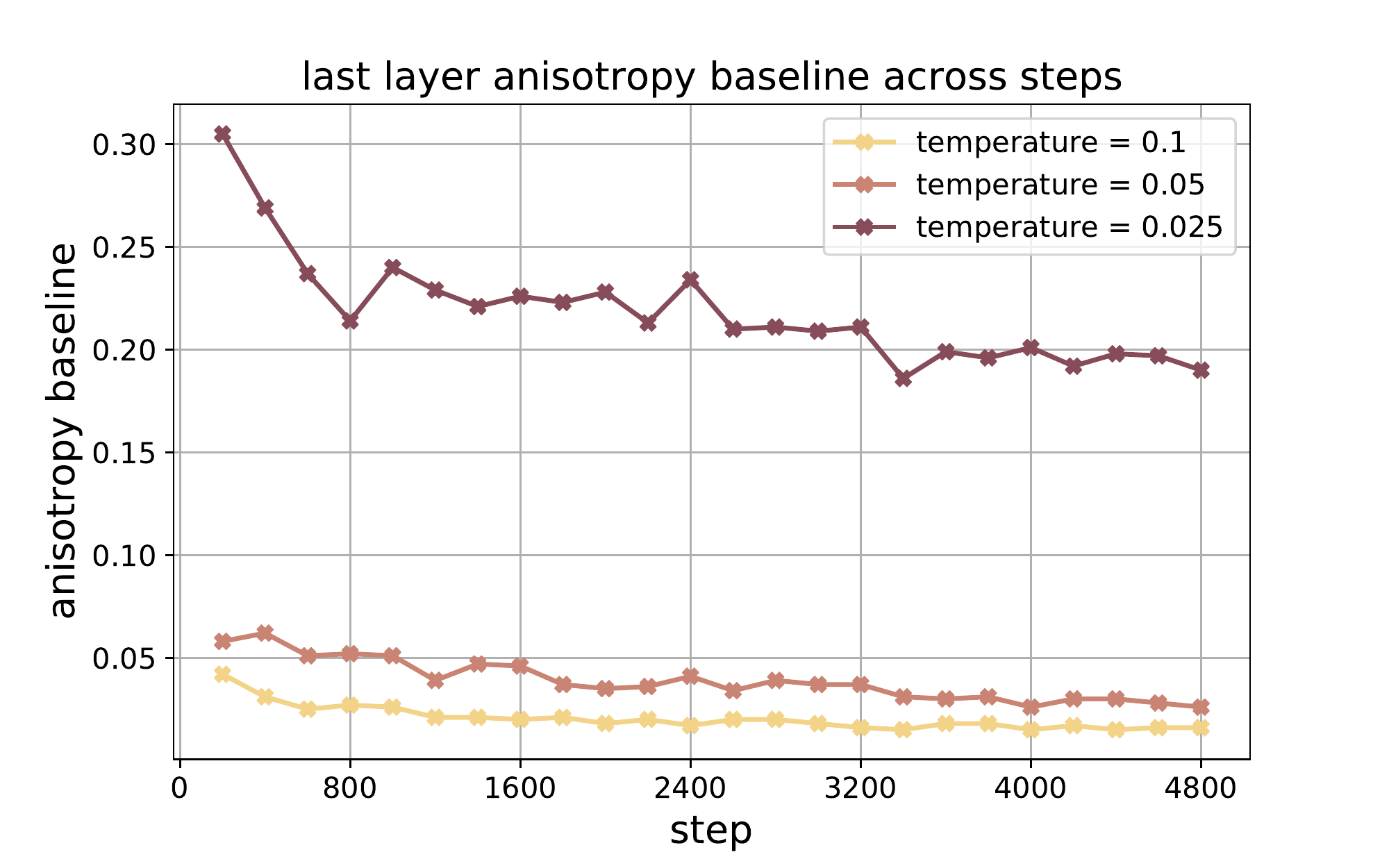}
\caption{Anisotropy changes throughout training under different temperatures} \label{temperature-anisotropy-baseline}
\end{figure}

\begin{figure}[h]
\includegraphics[width=7.7cm]{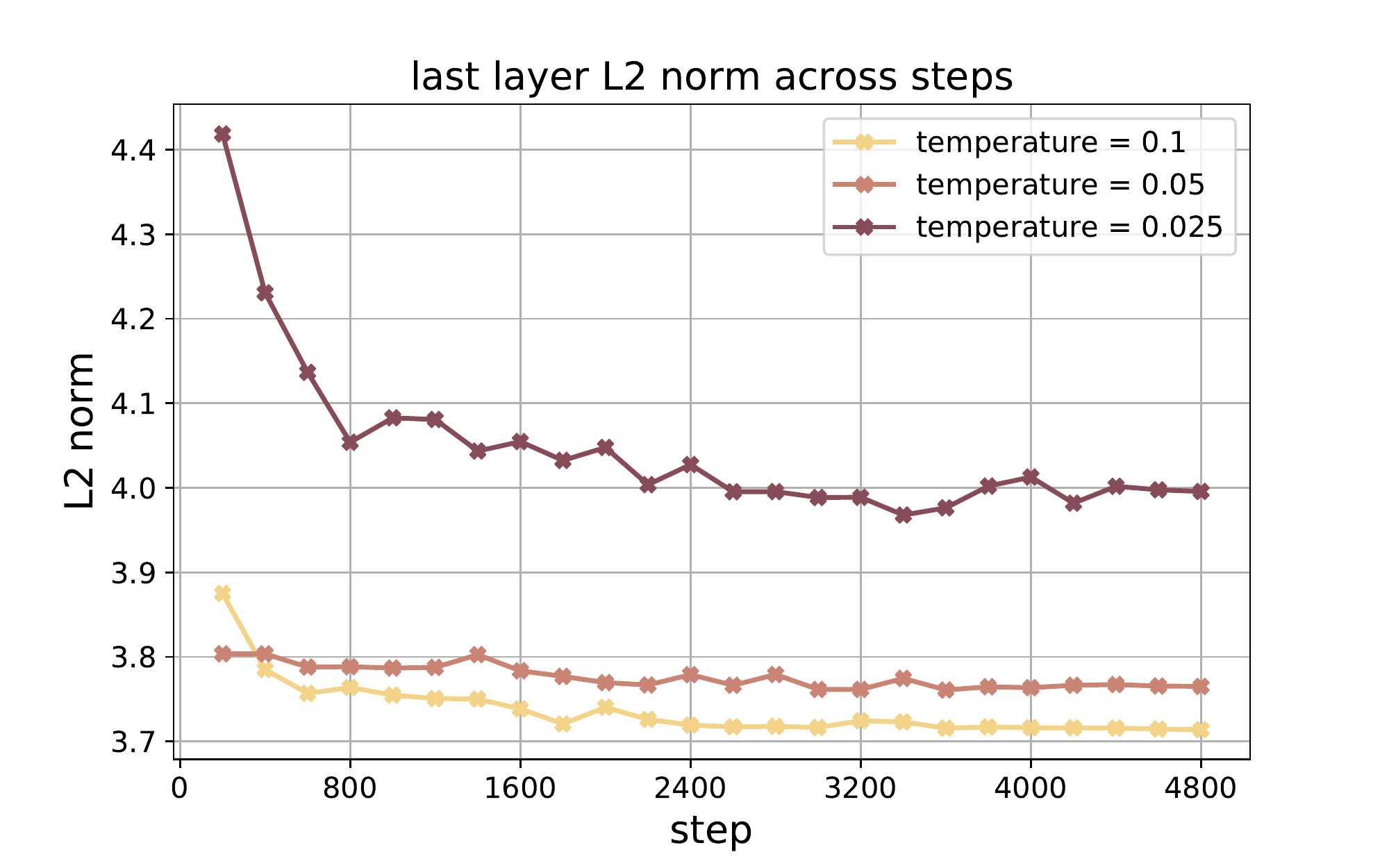}
\caption{L2-norm under different temperatures} \label{temperature-L2-norm}
\end{figure}
Firstly, we present the centered property we are measuring, anisotropy. Figure~\ref{temperature-anisotropy-baseline} shows the last-layer anisotropy change throughout steps. The trend is in line with our hypothesis about temperature being a "thurst". Knowing that the vanilla model starts from encoding embeddings to be stuck in a narrow angle, temperature serves as the power to push them further through forcing negative pairs to be different. With a higher temperature, the cosine similarity between negative pairs has to be lower to reach a similar loss. Figure~\ref{temperature-L2-norm} further validates this through showing that higher temperatures compress the semantic space in general, pushing instances to the origin.

\begin{figure}[h]
\includegraphics[width=7.7cm]{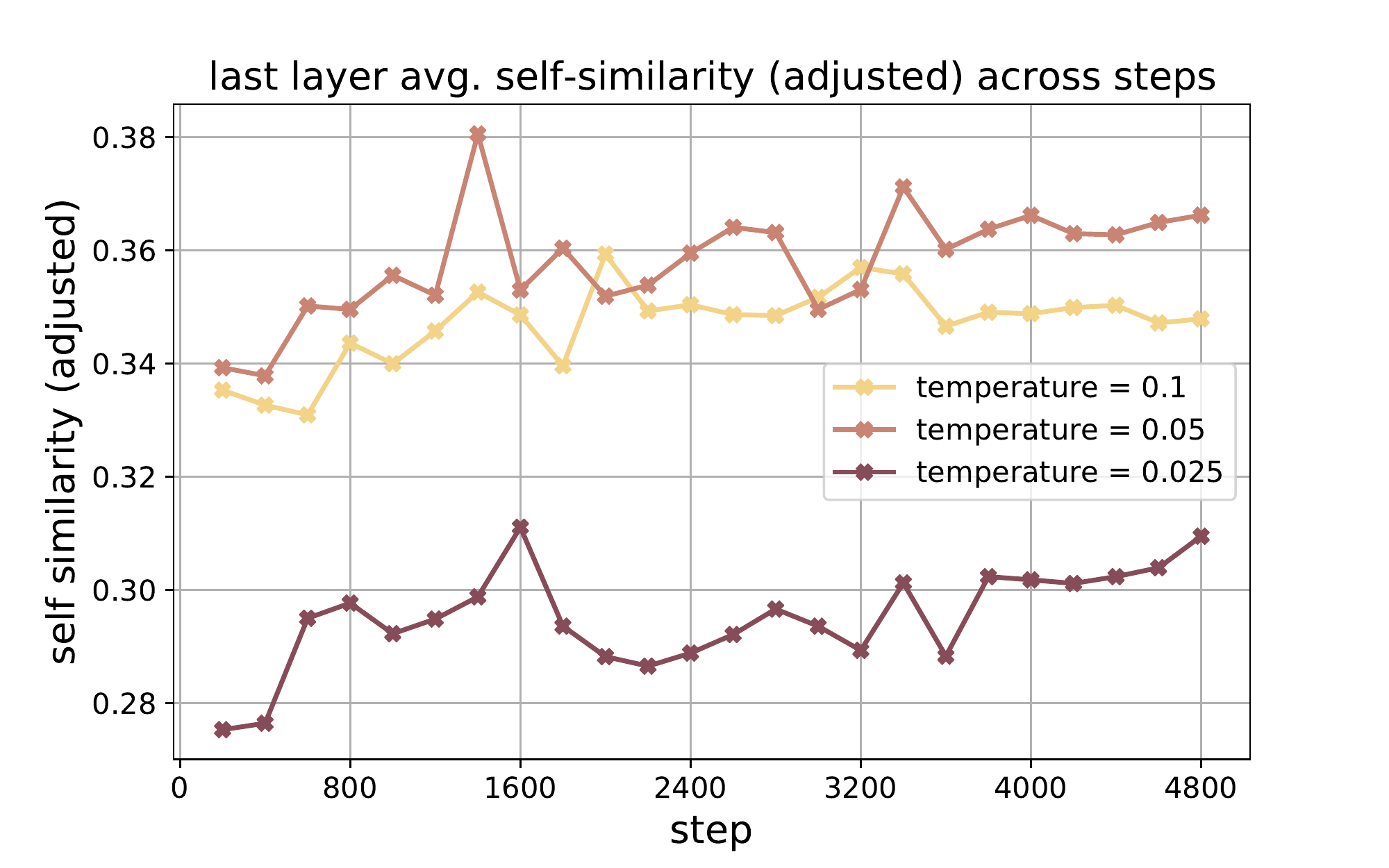}
\caption{Self similarity under different temperature, adjusted by anisotropy baseline} \label{temperature-self-adjusted}
\end{figure}

\begin{figure}[h]
\includegraphics[width=7.7cm]{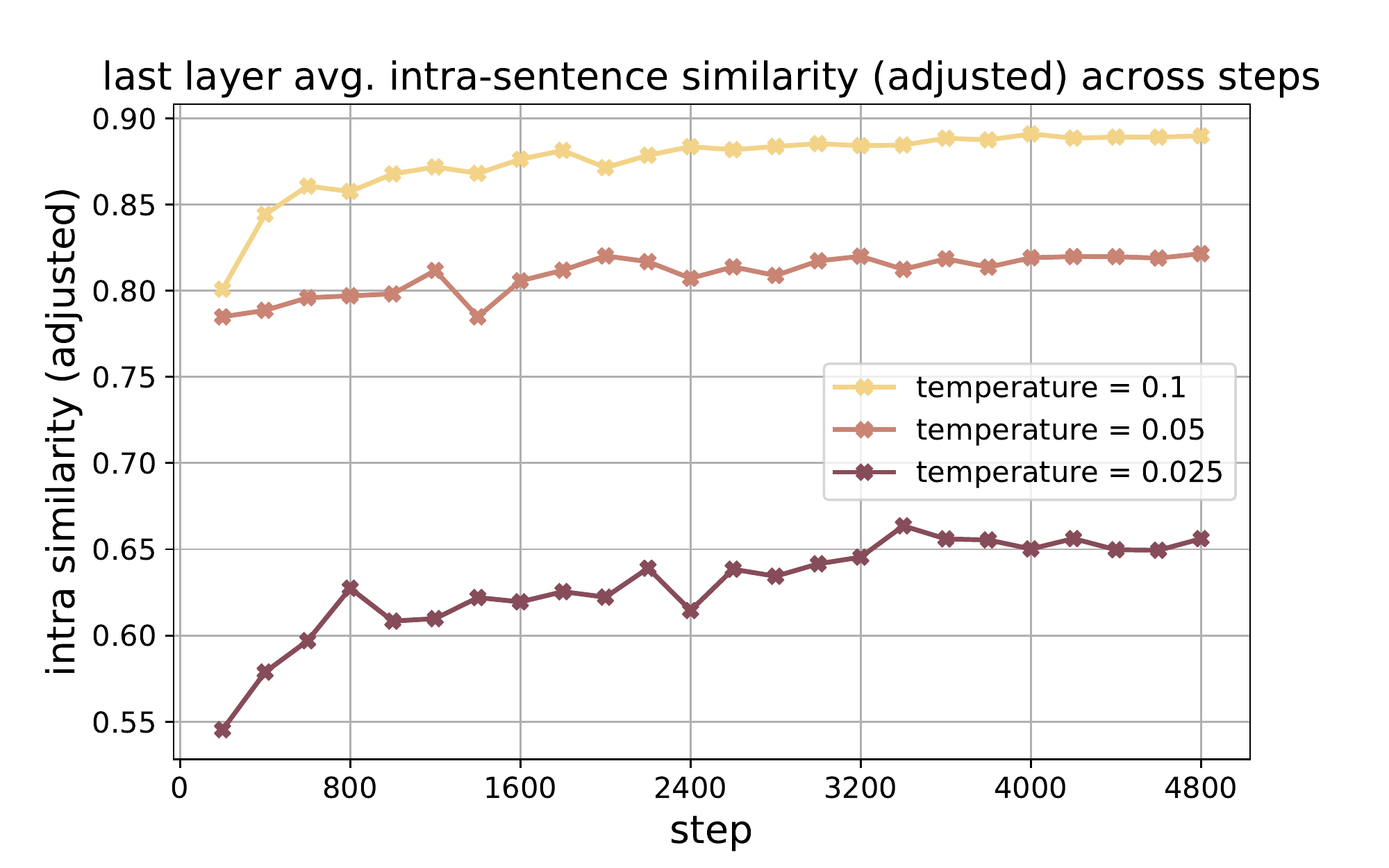}
\caption{Intra-sentence similarity under different temperature, adjusted by anisotropy baseline} \label{temperature-intra-adjusted}
\end{figure}
Figure~\ref{temperature-self-adjusted} and Figure~\ref{temperature-intra-adjusted} present the adjusted self and intra-sentence similarity. Following the closer look at the contradicted pattern for frequency bias analyzed in Section~\ref{sec:frequency bias}, the behavior here becomes self-explainable. We could see that under the temperature of 0.1, the self similarity stays at a lower level compared to 0.05 in the last steps. This matches with the opposite result in intra-sentence similarity. According to our analysis in Section~\ref{sec:frequency bias}, it is the less meaningful tokens that drag down the self-similarity, and because they learn to follow the semantically meaningful tokens wherever their embeddings go in the semantic space, the corresponding intra-sentence similairty would become much higher. We speculate that, while a high intra similarity explains the performance gain of models trained with contrastive loss on semantic tasks, its being too intense (as shown when $\tau$ = 0.1) might also account for the performance drop, making semantic meaningful tokens too dominating compared to auxiliary/functional tokens. Therefore, it again justifies the importance of selecting \textbf{a moderate temperature} that provides enough gradients, but not over-intensifying the attention leaning toward dominating tokens.

In Appendix~\ref{sec:batch size appendix}, we provide the analysis on batch size, revealing that batch size plays a less significant role, if given a relatively optimal temperature. This is the opposite of what is commonly found in visual representation learning. Appendix ~\ref{sec: pooling method} compares the three commonly used pooling methods, showing that the found patterns are not just artifacts of a certain pooling method (mean pooling), but consistent across pooling methods.

\section{Conclusion}

In this paper, we demystify the successes of using contrastive objectives for sentence representation learning through the lens of isotropy and learning dynamics. We showed the theoretical promise of uniformity brought by contrastive learning through measuring anisotropy, complemented by showing the flattened domination of top dimensions. We then uncovered a very interesting yet under-covered pattern: contrastive learning learns to converge tokens in a same sentence, bringing extremely high intra-sentence similarity. We then explained this pattern by connecting it to frequency bias, and showed that semantically functional tokens fall back to be the by products of semantically meaningful tokens in a sentence, following wherever they travel in the semantic space. Lastly, we ablate all findings through temperature, batch size and pooling method, providing a closer look at these patterns through different angles.

% We hope this research can inspire better design of contrastive learning for sentence representation learning in different domains and different settings.

\section{Limitations}

This paper only considers analyzing contrastive learning in the fine-tuning stage, but we note that with isotropy being a desiderata for pre-trained language models \cite{ethayarajh2019contextual}, recent works have considered incorporating contrastive objectives in the pre-training stage \cite{izacardunsupervised,su-etal-2022-tacl}. We leave analysis on this line of research for future work. 

We further note that the analysis in this work focuses on theoretical properties occurred during contrastive SRL (e.g., high intra-sentence similarity), thus only focuses on semantic textual similarity (STS) data as a proof of concept. However, with the growing attention on contrastive learning, we argue that the typical STS-B is perhaps no longer sufficient for revealing the full ability of models trained with newer contrastive SRL frameworks. We call for a standard practice that the performance of contrastive SRL should be assessed on both semantic textual similarity and information retrieval tasks (e.g., \citet{thakur2021beir}). We leave analysis on information retrieval tasks leveraging our analysis pipeline for future studies. For example, how high intra-sentence similarity is related to the learned attention towards tokens that enable document retrieval with better performance.

\bibliography{anthology,custom}
\bibliographystyle{acl_natbib}
\appendix

\section{Top Self Similarity Change (SSC): Token Examples}

Table~\ref{frequent words self-sim change} presents top 10 positive and negative self similarity change of frequent tokens, before and after contrastive fine-tuning. 

Although function tokens are found to be highly contextualized in pre-trained language models \cite{ethayarajh2019contextual}, this phenomenon is even intensified after contrastive fine-tuning. While for semantic tokens, the spurious contextualization is alleviated to a great extent.

\section{Expanded semantic space (Eased Anisotropy)}

We provide a visualization of embedding geometry change in Figure~\ref{expanded semantic space}.

We first use the vanilla mpnet to encode the STS-B subset we have constructed. During fine-tuning, we save the models every 200 steps and use them to encode the subset, We find that with optimal hyperparameters, the representations go through less change after 200 steps. We perform UMAP dimensionality reduction on embeddings provided by models up to 1000 step to preserve better global structure, and visualize only vanilla and 200-step embeddings.

\section{Unadjusted measures of Section~\ref{ots results}}
\label{sec:unadjusted measure appendix}

\begin{figure}[h]
\includegraphics[width=7.7cm]{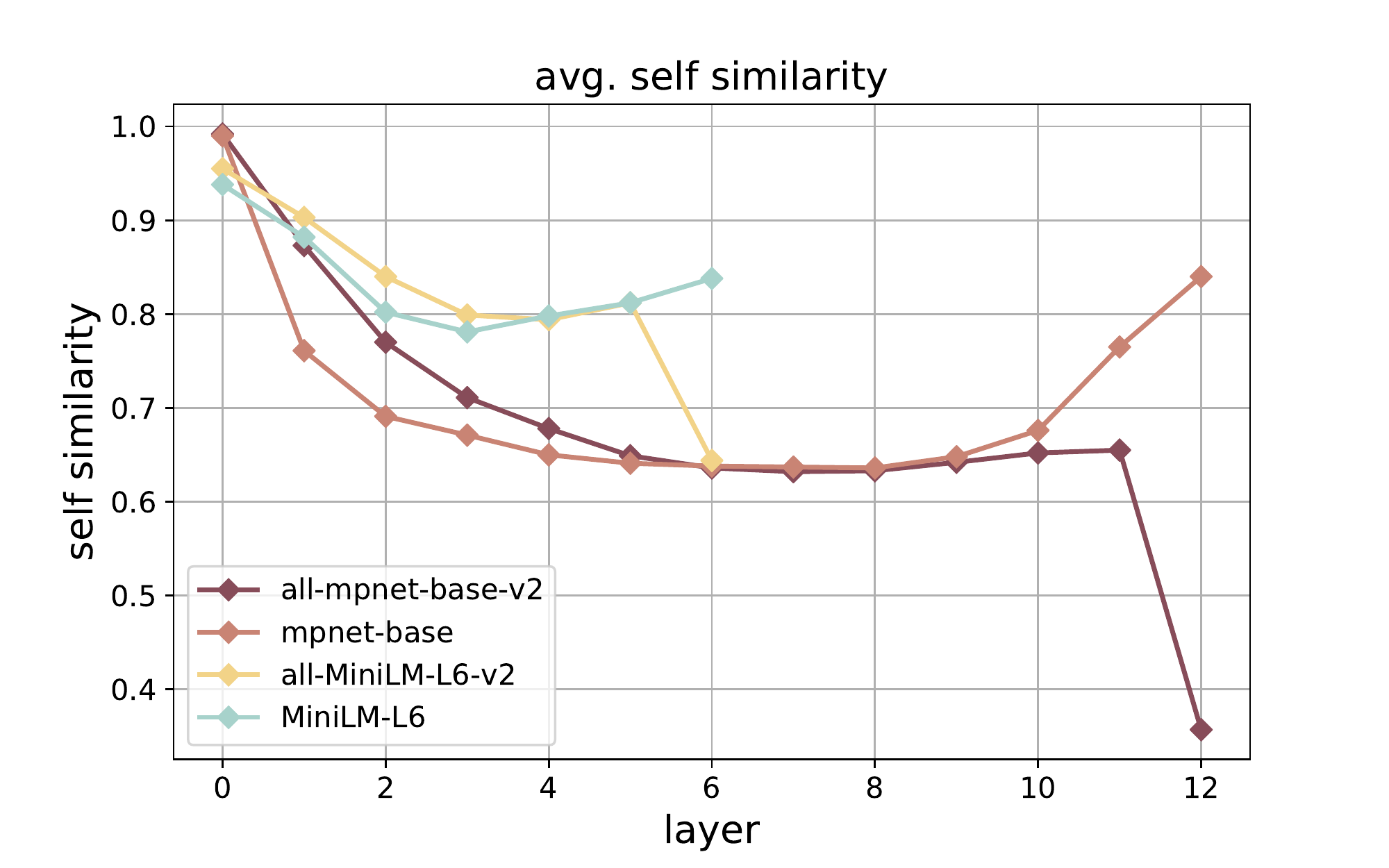}
\caption{Unadjusted self similarity of tokens} \label{ots-self-sim}
\end{figure}

\begin{figure}[h]
\includegraphics[width=7.7cm]{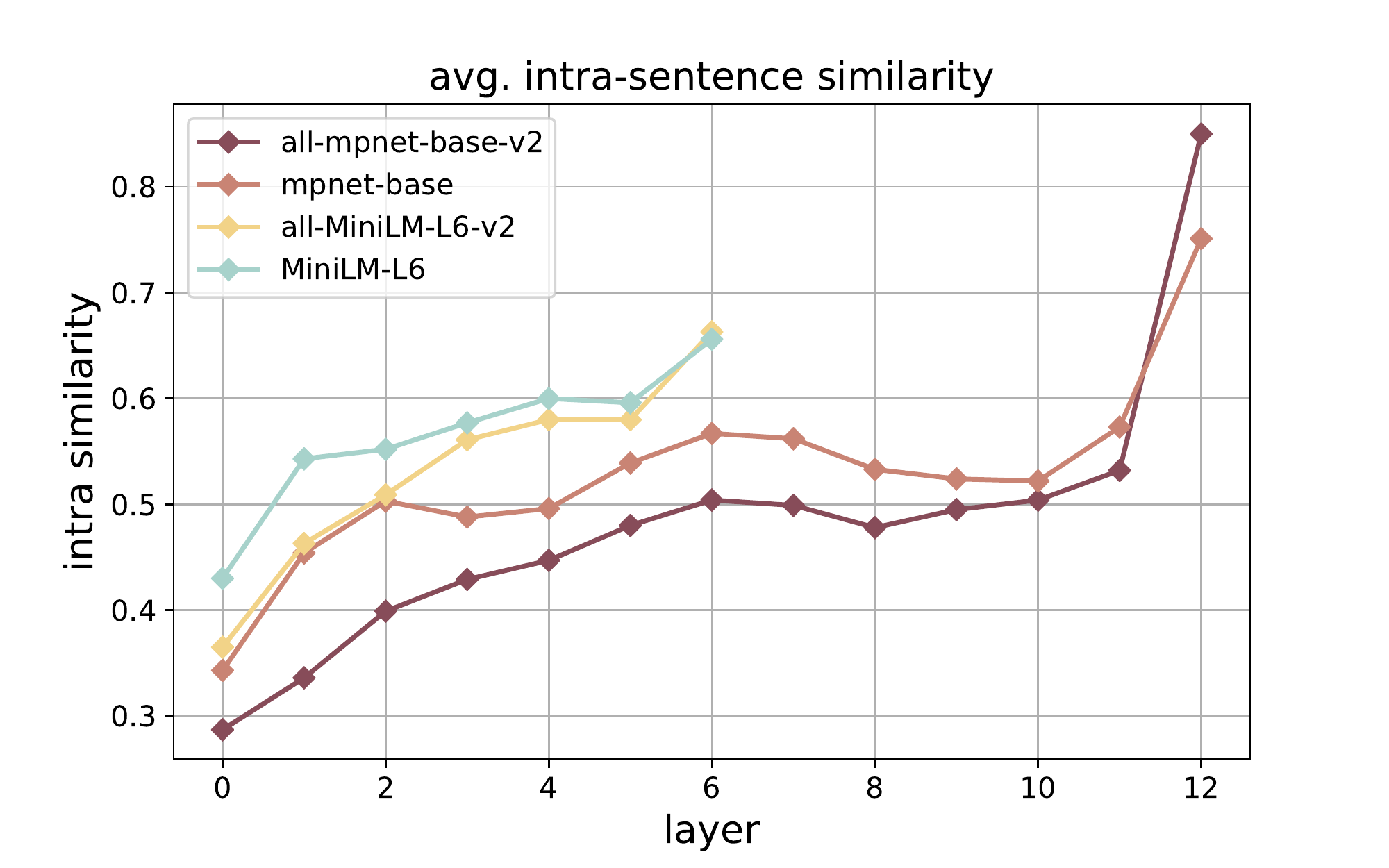}
\caption{Unadjusted intra-sentence similarity of tokens} \label{ots-intra-sim}
\end{figure}

Figure~\ref{ots-self-sim} and Figure~\ref{ots-intra-sim} display respectively the unadjusted avg. self similarity and intra-sentence similarity. These values as we elucidated in previous sections, however, are likely to be artifact of anisotropy, and therefore are supposed to be adjusted by the anisotropy baseline of each model, based on the computation on randomly sampled token pairs.

As shown in main sections, to offset the effect of each model's intrinsic non-uniformity, we adjust them by the degree of anisotropy of each model, based on pair-wise average similarity among 1000 token representations that we randomly sample from each of the 1000 sentences (to avoid the sampling to bias toward long sentences). 

\section{Temperature Search: why searching to the order of magnitude by 10 is not optimal?}
\label{sec:temperature search appendix}

We have also run the search range of temperature in previous research, which is carried out to the order of magnitude by 10. We compare the metrics on the models run with these temperatures with the vanilla mpnet model's performance.

It is shown that, not all values of temperature push the metrics from the vanilla baseline toward a same direction. Therefore, there exists a relatively optimal range to search, which is empirically implemented in a few works \cite{yan2021consert,zhang2022contrastive}, but few seems to have discussed why the range should not be that large, while we show this through the math analysis in Section~\ref{section: tempearature} and their contradicted performance on our studied metrics here.

Specifically, for anisotropy baseline, temperature being too low even augments the vanilla model's unideal behavior, and the same applies for L2-norm, by that temperature being too low actually pushes the embeddings even further from the origin.
\begin{figure}[h]
\includegraphics[width=7.7cm]{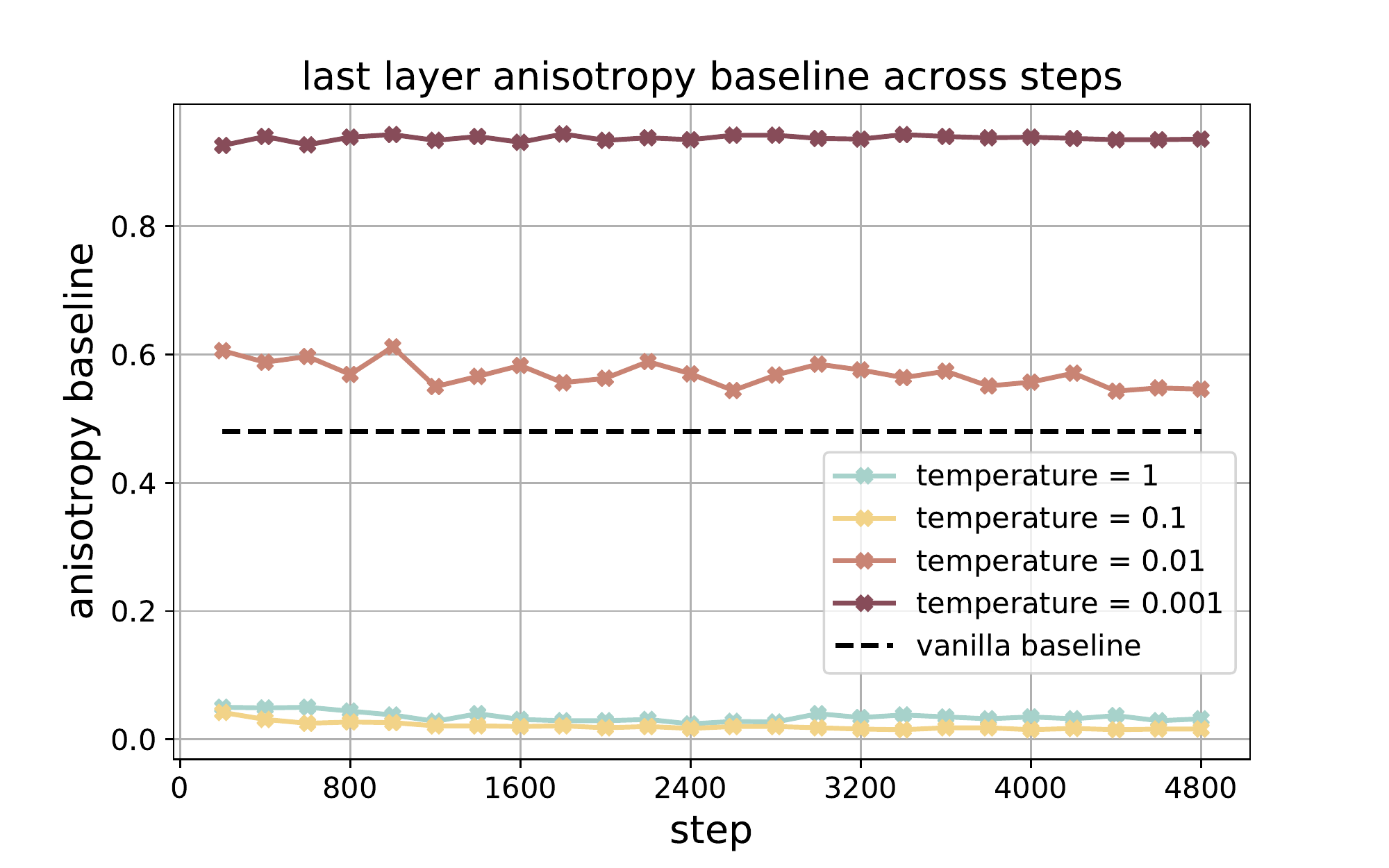}
\caption{Anisotropy changes throughout training under different temperatures} \label{temperature-anisotropy-baseline-search10}
\end{figure}

\begin{figure}[h]
\includegraphics[width=7.7cm]{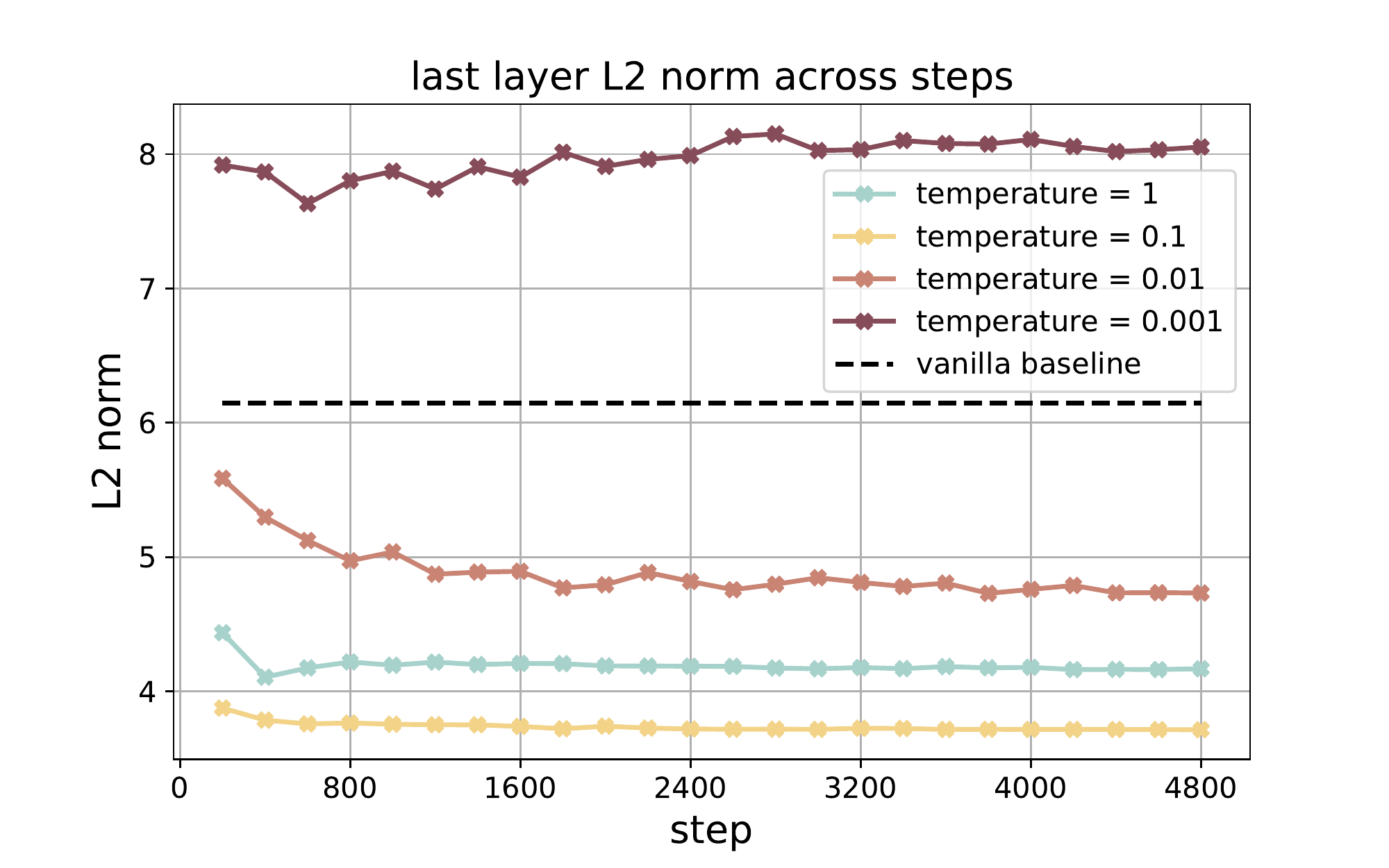}
\caption{L2-norm under different temperatures} \label{temperature-L2-norm-search10}
\end{figure}
For the adjusted self similarity and intra-sentence similarity, the metrics for low temperature are largely offset by anisotropy, meaning that for these temperature (especially $\tau = 0.001$), tokens are not more similar to itself in different contexts, nor to other tokens they share contexts with, compared to just with a random token in whatever context.

\begin{figure}[h]
\includegraphics[width=7.7cm]{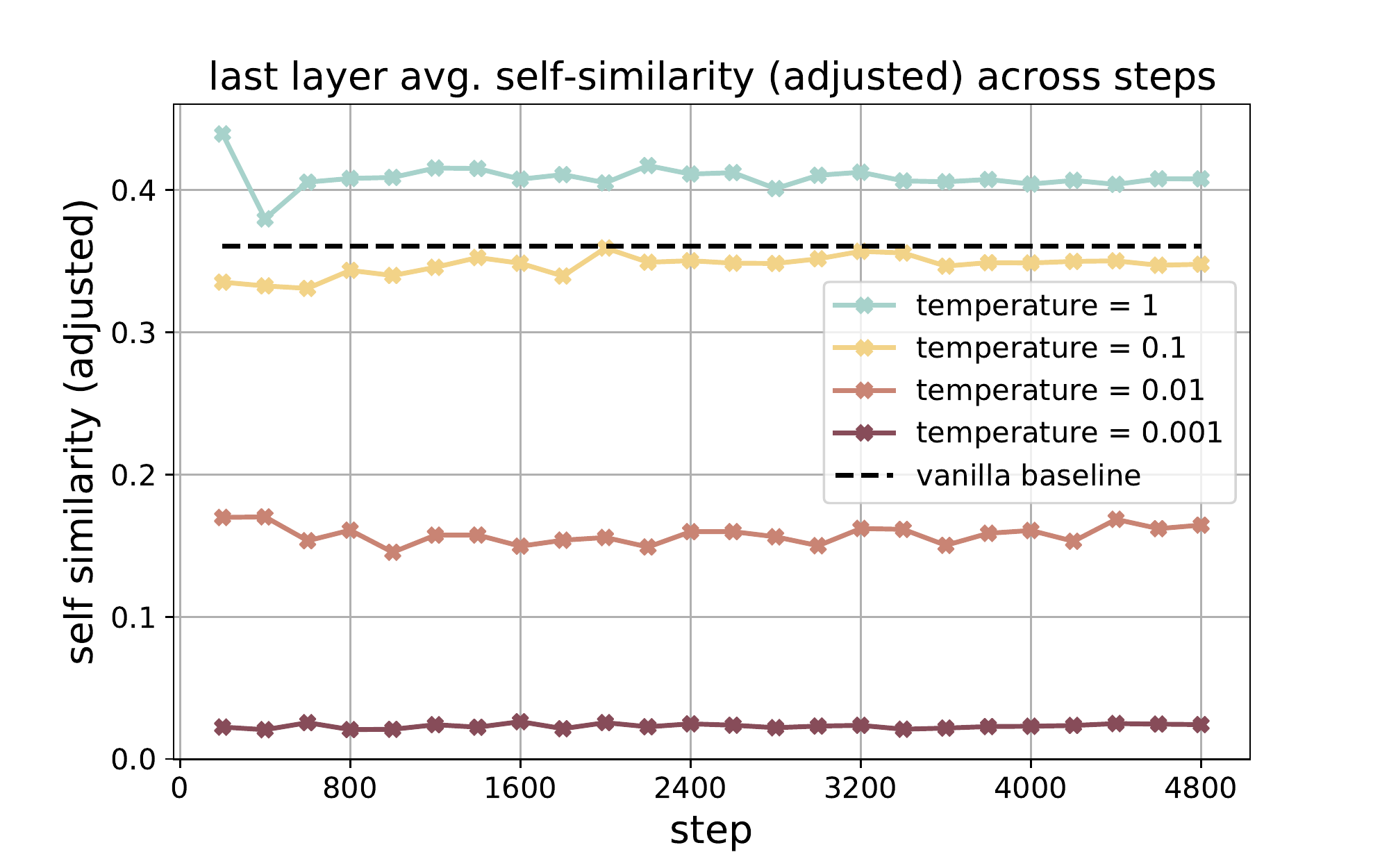}
\caption{Self similarity under different temperatures, adjusted by anisotropy baseline} \label{temperature-self-adjusted-search10}
\end{figure}

\begin{figure}[h]
\includegraphics[width=7.7cm]{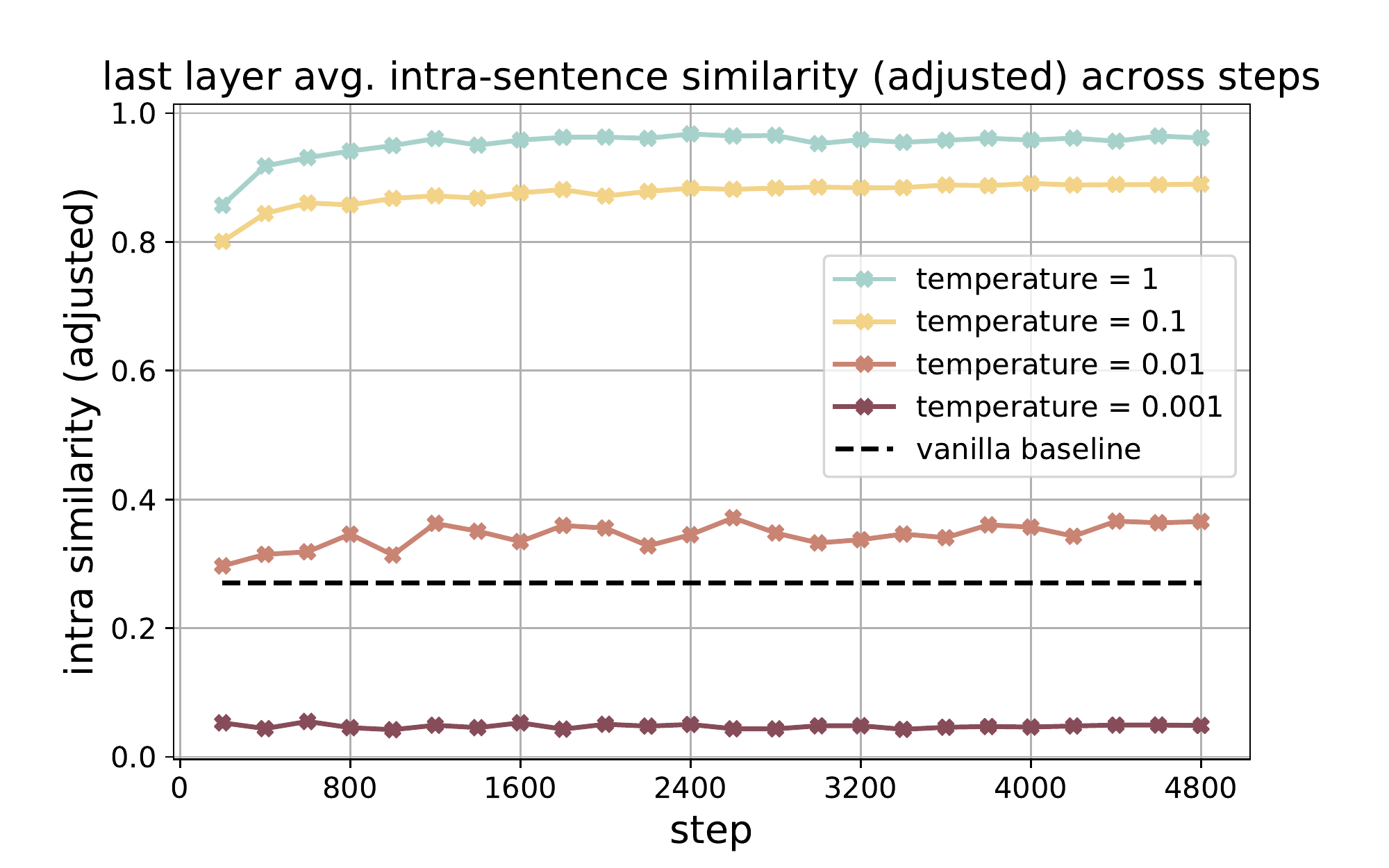}
\caption{Intra-sentence similarity under different temperatures, adjusted by anisotropy baseline} \label{temperature-intra-adjusted-search10}
\end{figure}

\citet{gao2018representation} and \citet{gao2021simcse} take a singular spectrum perspective in understanding regularization to anisotropy. \citet{gao2018representation} propose a regularization term to the original log-likelihood loss in training machine translation model to mitigate the representation degeneration problem (or anisotropy). The regularization is proportional to $Sum(WW^T)$
, where $W$ is the stack of normalized word embeddings. If all elements are positive, then minimizing $Sum(WW^T)$ is equivalent to minimizing the upper bound for the largest top eigenvalue of $Sum(WW^T)$. Therefore, this regularization term shows theoretical promise to flatten the singular spectrum and make the representation more uniformly distributed around the origin. \citet{gao2021simcse} extend this analysis to show the same theoretical promise brought by the uniformity loss proposed by \citet{wang2020understanding}, by deriving that uniformity loss is in fact greater or equal to $\frac{1}{\tau m^2}\sum\limits_{i=1}^m\sum\limits_{j=1}^m h_i^Th_j$, which is also equivalent to flattening the spectrum of the similarity matrix. Our results show that despite the intuition reached by singular spectrum perspective, the assumption could probably only hold on a relatively optimal temperature. Thus, the effect of temperature should be considered using this perspective, which is beyond the scope of this paper.

\section{Batch size}
\label{sec:batch size appendix}

Batch size on the other hand, does not produce impact as significant as temperature. We have run three models with the optimal $\tau = 0.05$ paired with a batch size range of $\{16, 64, 256\}$. 

The metrics yielded by different batch sizes all stay in small range at the end of the epoch, albeit showing different rates and stability of convergence.

\begin{figure}[h]
\includegraphics[width=7.7cm]{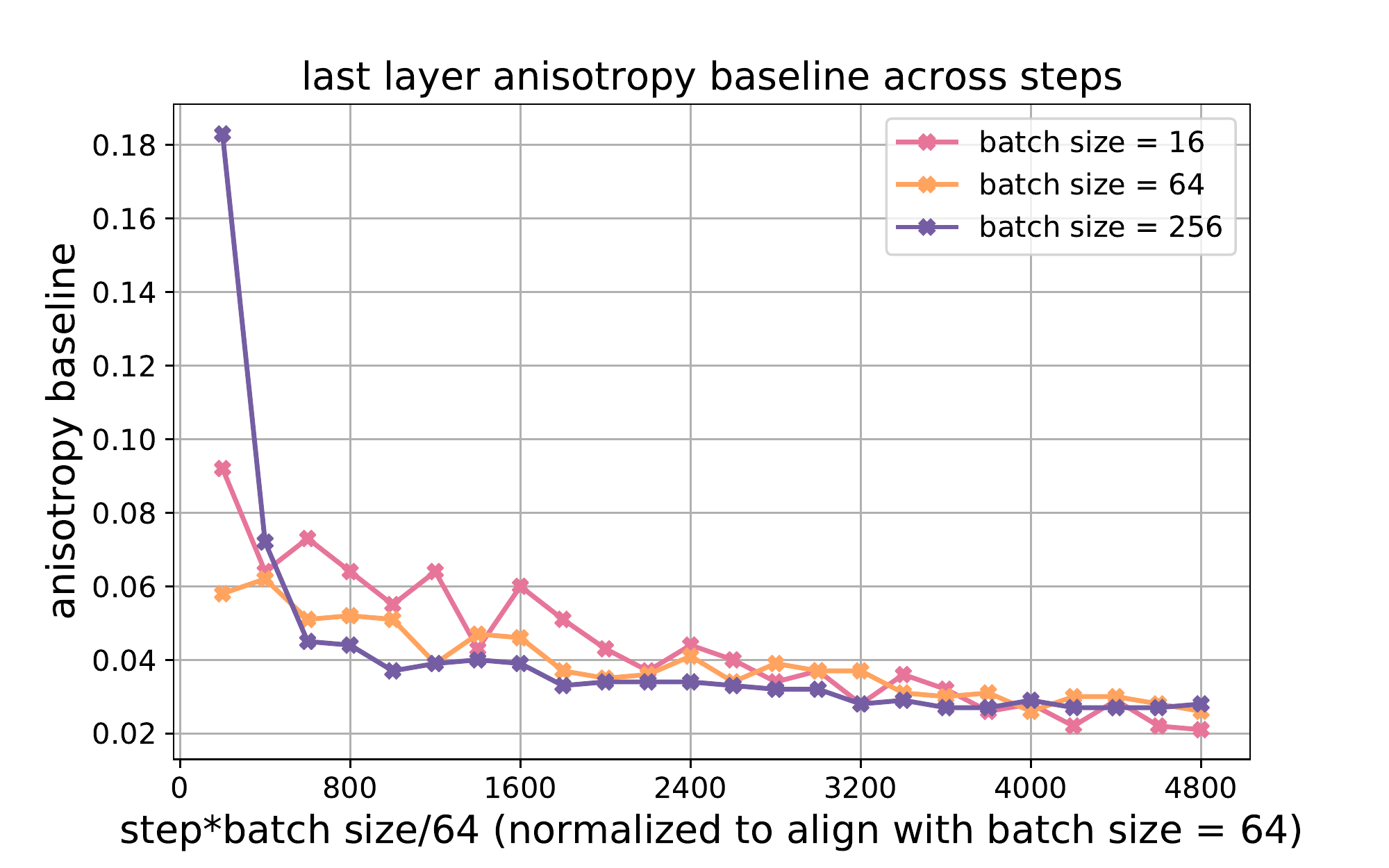}
\caption{Anisotropy changes throughout training under different batch sizes} \label{batch-anisotropy-baseline}
\end{figure}

\begin{figure}[h]
\includegraphics[width=7.7cm]{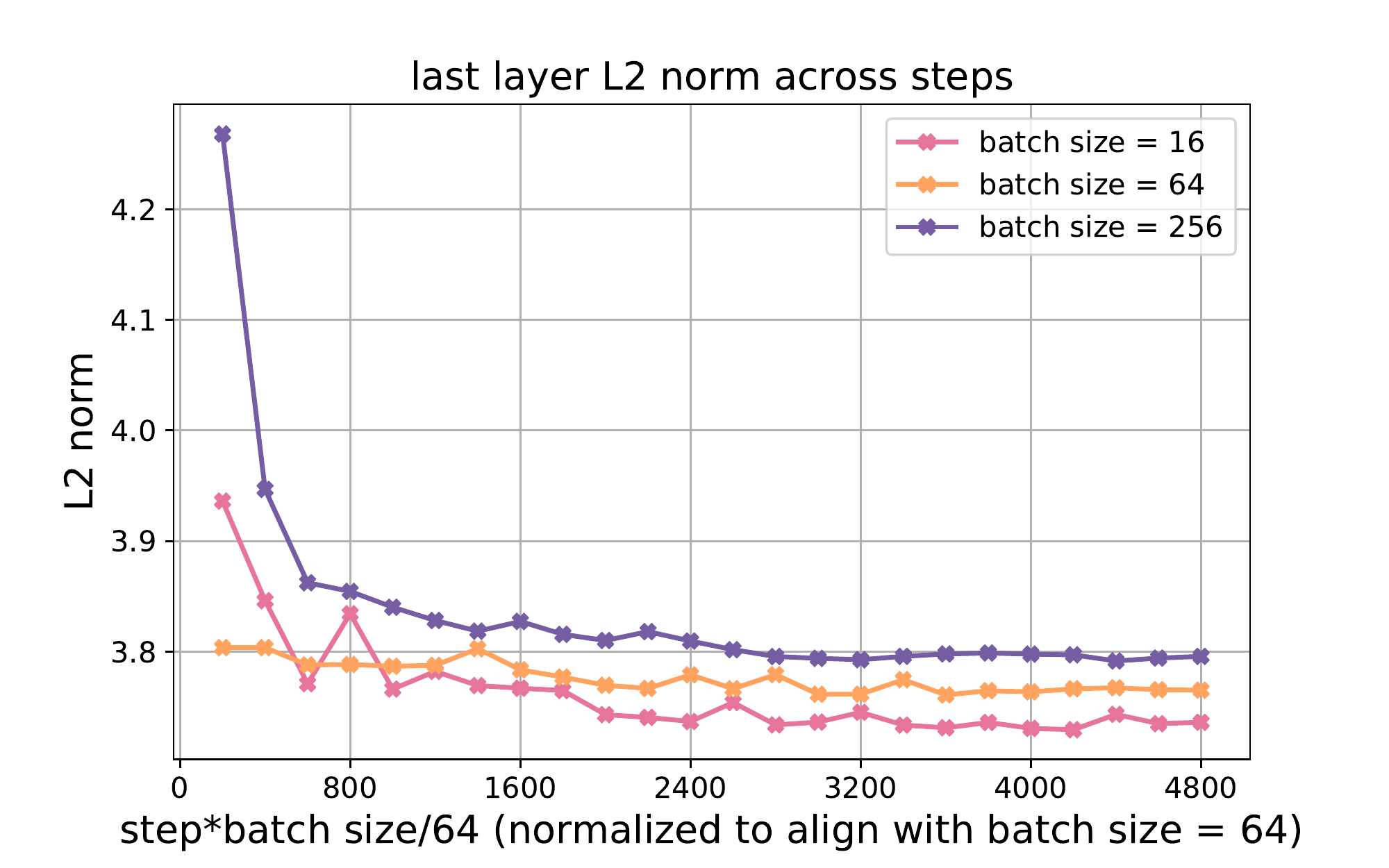}
\caption{L2-norm under different batch sizes} \label{batch-L2-norm}
\end{figure}

\begin{figure}[h]
\includegraphics[width=7.7cm]{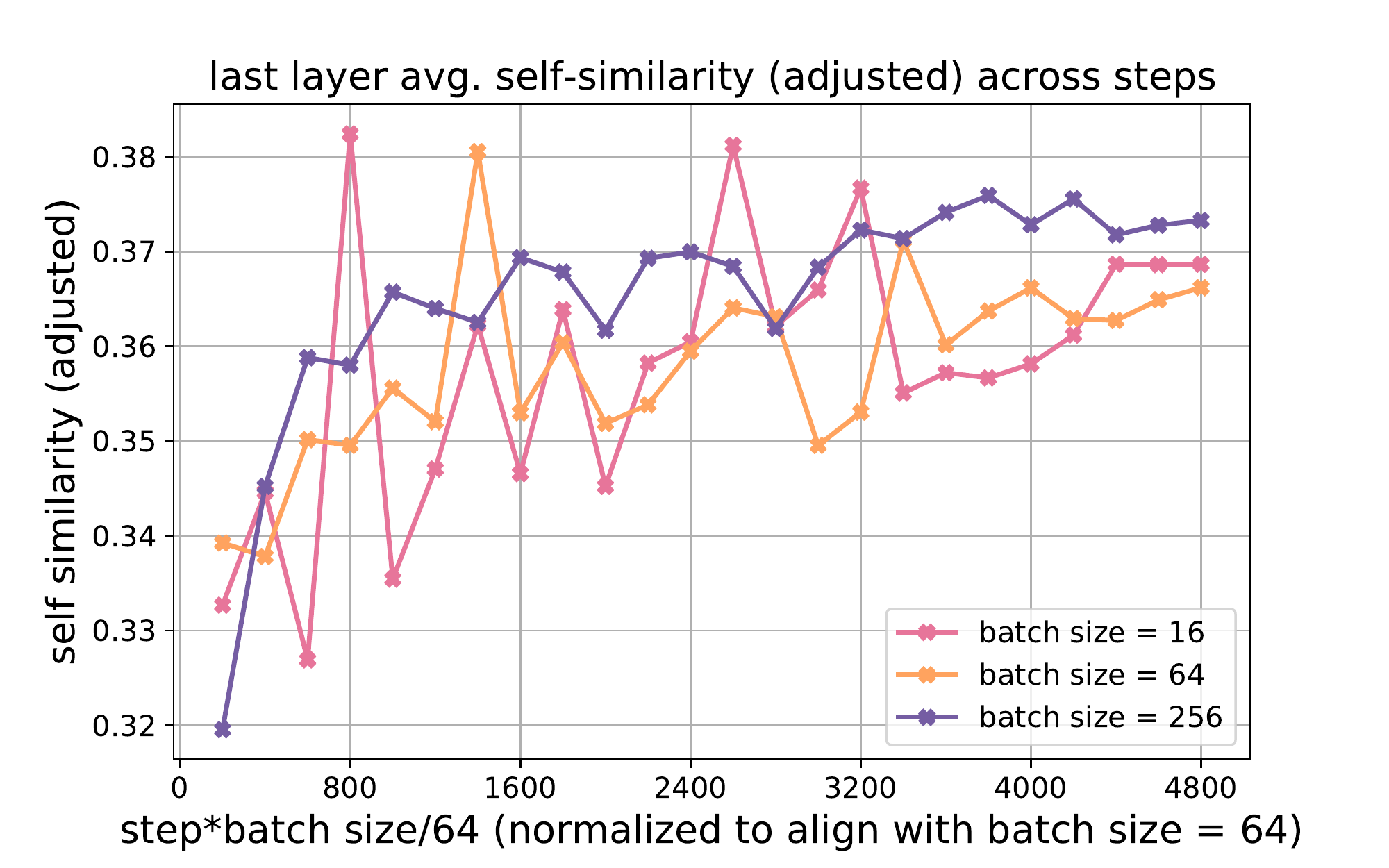}
\caption{Self similarity under different batch sizes, adjusted by anisotropy baseline} \label{batch-self-adjusted}
\end{figure}

\begin{figure}[h]
\includegraphics[width=7.7cm]{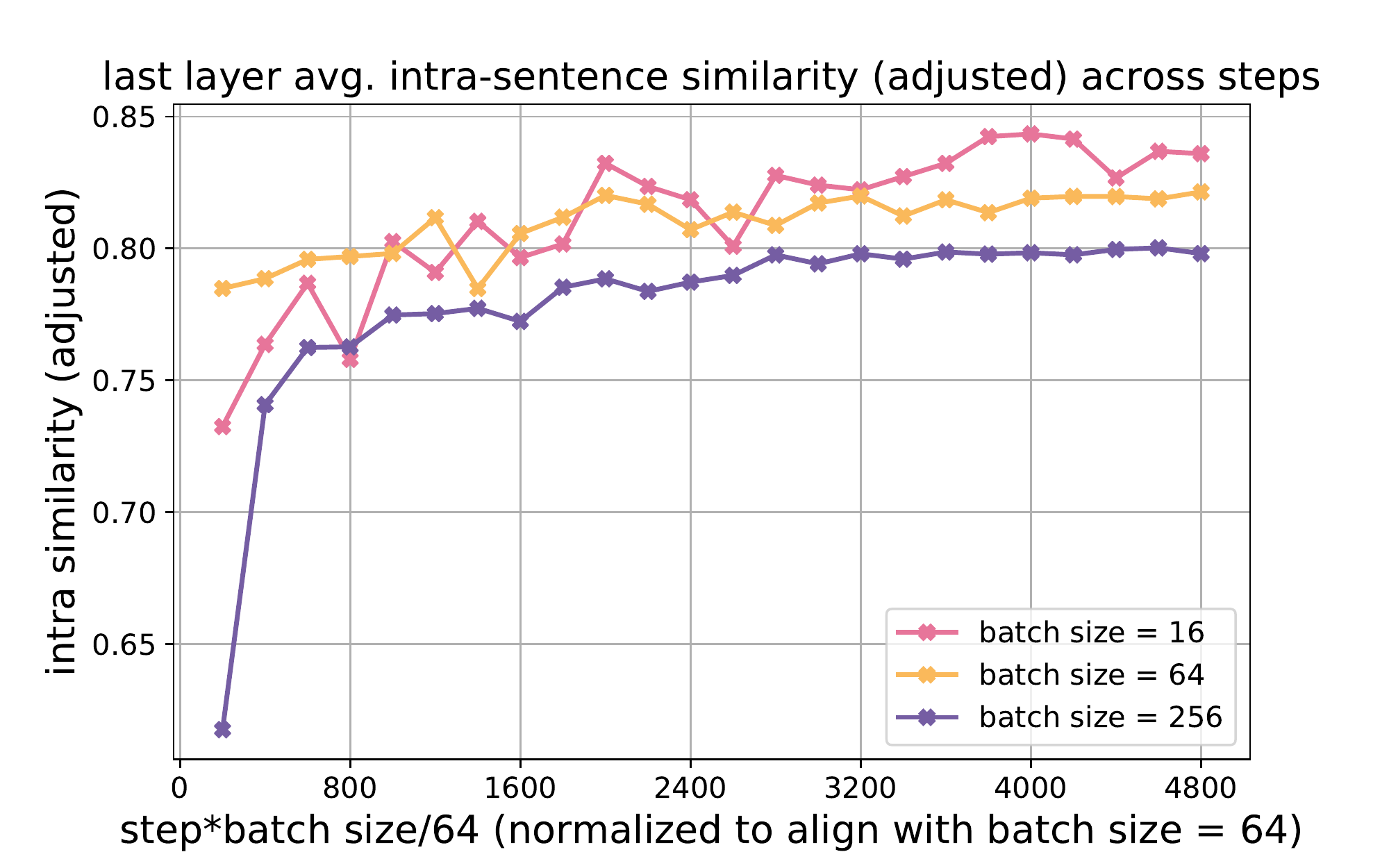}
\caption{Intra-sentence similarity under different batch sizes, adjusted by anisotropy baseline} \label{batch-intra-adjusted}
\end{figure}

\begin{table*}[h]
\centering
\resizebox{15cm}{!}{
\begin{tabular}{lllllllllll}
\hline
\textbf{Model} & \textbf{k = 1} & \textbf{k = 2} & \textbf{k = 3} &\textbf{k = 5}&\textbf{k = 10}&\textbf{k = 20}&\textbf{k = 50}&\textbf{k = 100}&\textbf{k = 300}&\textbf{k = 700}\\
\hline
\hline
mpnet\textsubscript{vanilla} & .386  & .338 & .210 & .169  & .168 & .182 & .201  & .195 & .175 & .040\\
mpnet\textsubscript{fine-tuned} & \textbf{.999}  & \textbf{.998} & \textbf{.996} & \textbf{.994}  & \textbf{.990} & \textbf{.983} & \textbf{.960}  & \textbf{.922} & \textbf{.783} & \textbf{.229}\\
\hline
minilm\textsubscript{vanilla} & .993  & \textbf{.980} & \textbf{.970} & \textbf{.947} & \textbf{.886} & .796 & .559  & .543 & \textbf{.375} & / \\
minilm\textsubscript{fine-tuned} & \textbf{.998}  & .846 & .836 & .830  & .817 & \textbf{.805} & \textbf{.768}  & \textbf{.690} & .285& /\\
\hline

\end{tabular}
}
\caption{\label{informativity table} $r^2$ between the similarity matrices of sampled token embeddings, before and after removing the same top-k rogue dimensions from every token embedding.}
\end{table*}

\section{Informativity}

\label{informativity}

In this section we present the informativity analysis outlined in Section~\ref{section: isotropy analysis}. Specifically, after we identify how dominant are the top rogue dimensions, to what degree is semantics affected with these rogue dimensions removed? Do these dimensions only have large mean but do not contribute to large variance? We sample 1k token embeddings to compute their pair-wise similarity. After removing top-k dimensions from every embedding, we compute the similarity matrix again, and compute the Pearson Correlation $r$ between flattened lower triangles of the matrices of the two excluding their diagonals. We then report the $r^2$ which represents the proportion of variance in the original similarity matrix explained by the post-processed matrix.

At a high level, Table~\ref{informativity table} shows that dominance $\neq$ informativity. Specifically, MiniLM presents a misalignment between dominance toward similarity computation and the actual information stored in these dimensions. For instance, removing the top 1 dominant dimension of minilm\textsubscript{finetuned} seems to not affect the embeddings' relative similarity to one another at all, preserving an $r^2$ of $.998$. Also, recall from Section~\ref{section: isotropy analysis} that contributions of dimensions from minilm\textsubscript{vanilla} to similarity computation are relatively flatter than mpnet\textsubscript{vanilla}, the results show that along with the even more flattened contributions after fine-tuning, the informativity seems to have been reallocated. For instance, from removing $k = 100$ to $k = 300$, the explainable variance goes down from $.690$ to $.285$, meaning this range of dimensions store a lot more information compared with the vanilla version. In general, that minilm\textsubscript{vanilla} and minilm\textsubscript{fine-tuned} take turn to yield higher $r^2$ with top-k removed demonstrates that there is generally no strong correlation between dominance and informativity, but it is rather random - especially when the dominance is already quite evenly distributed in the vanilla model.

\section{Pooling Method}
\label{sec: pooling method}
In line with previous analysis, this section presents the measurement on different pooling methods. We follow the same setting in Section~\ref{section: tempearature} to also investigate whether the patterns found in Section~\ref{section: isotropy analysis} are only attributable to mean pooling. We compare mean pooling with [cls] pooling and max pooling. Albeit the different performance on the metrics, contrastive learning in general presents consistent behaviors across pooling methods, such as eased anisotropy and enhanced intra-sentence similarity For anisotropy, we observe that [cls] pooling shows a slow convergence on producing isotropy. At the end of the epoch, it is still on a decreasing trend. By contrast, mean pooling and max pooling demonstrate a faster convergence, with mean pooling being most promising on isotropy. Their performance on L2-norm is also well-aligned, again showing strong correlation between isotropy and L2-norm in the training process utilizing contrastive loss. And this correlation seems agnostic to pooling methods. The following analysis focuses on their differences:
\begin{figure}[h]
\includegraphics[width=7.7cm]{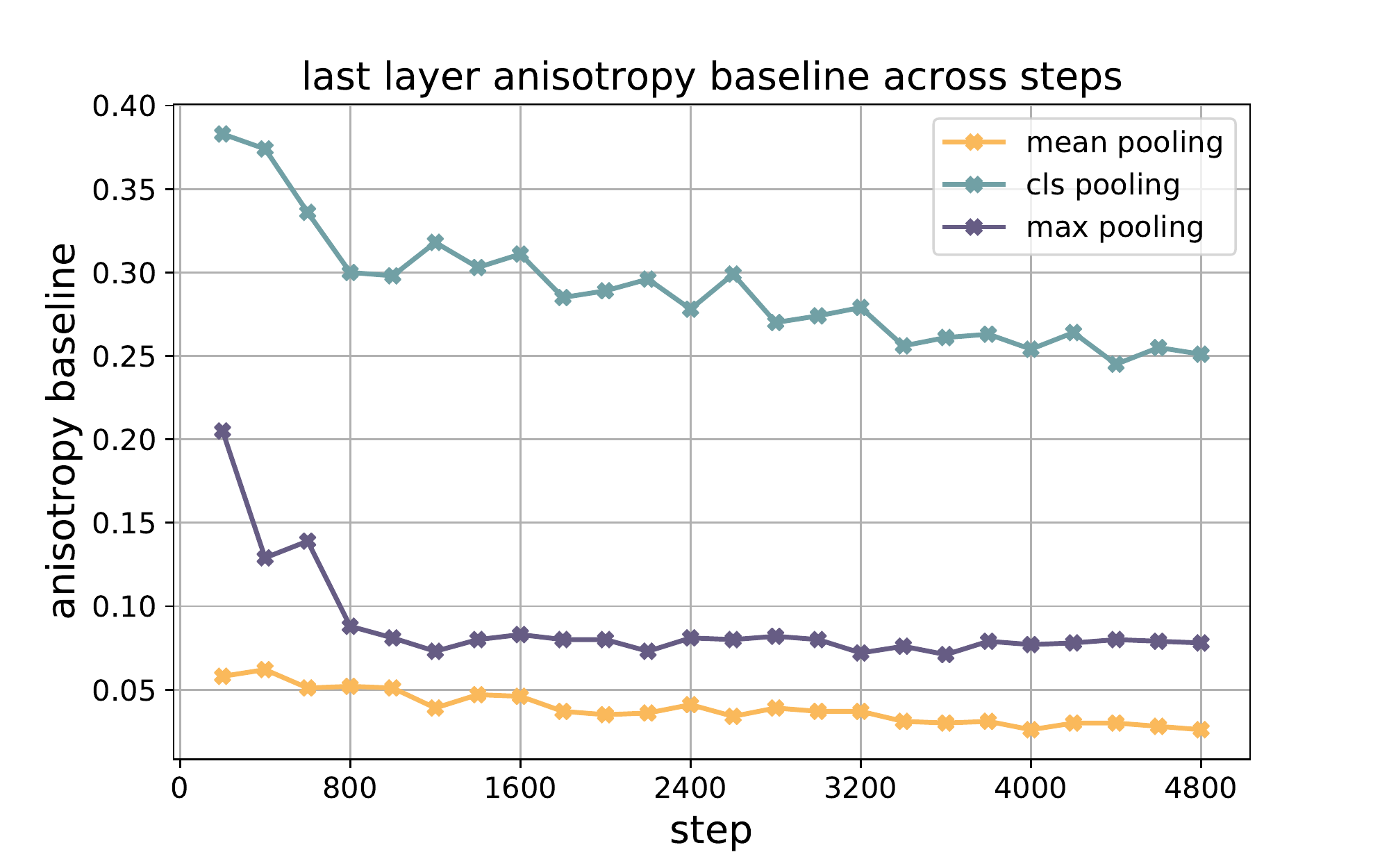}
\caption{Anisotropy changes throughout training under different pooling methods} \label{temperature-anisotropy-baseline-pooling}
\end{figure}

\begin{figure}[h]
\includegraphics[width=7.7cm]{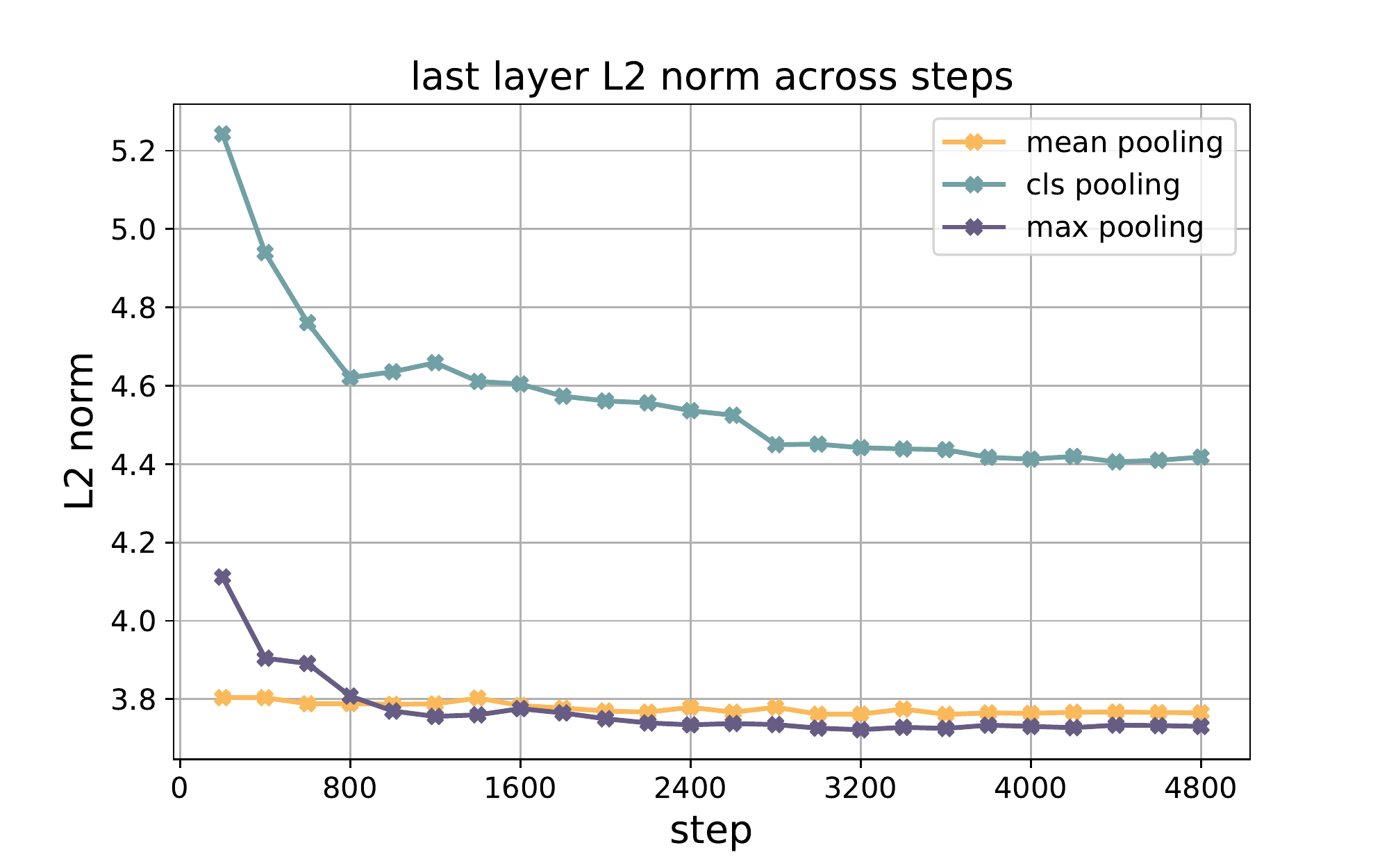}
\caption{L2-norm under different pooling methods} \label{temperature-L2-norm-pooling}
\end{figure}
For self similarity, [cls] pooling and mean pooling show a similar performance, which max pooling deviates from. 

Max pooling presents an "unacceptably" high intra-sentence similarity. Although intra-sentence similarity is a potentially ideal property uniquely brought by contrastive learning, this metric could not be over-intensified, as also shown in Section~\ref{section: tempearature}, Appendix~\ref{sec:temperature search appendix}, and Appendix~\ref{sec:batch size appendix}. There exists an ideal range for intra-sentence similarity, compatible to a model's performance on other metrics.

\begin{figure}[h]
\includegraphics[width=7.7cm]{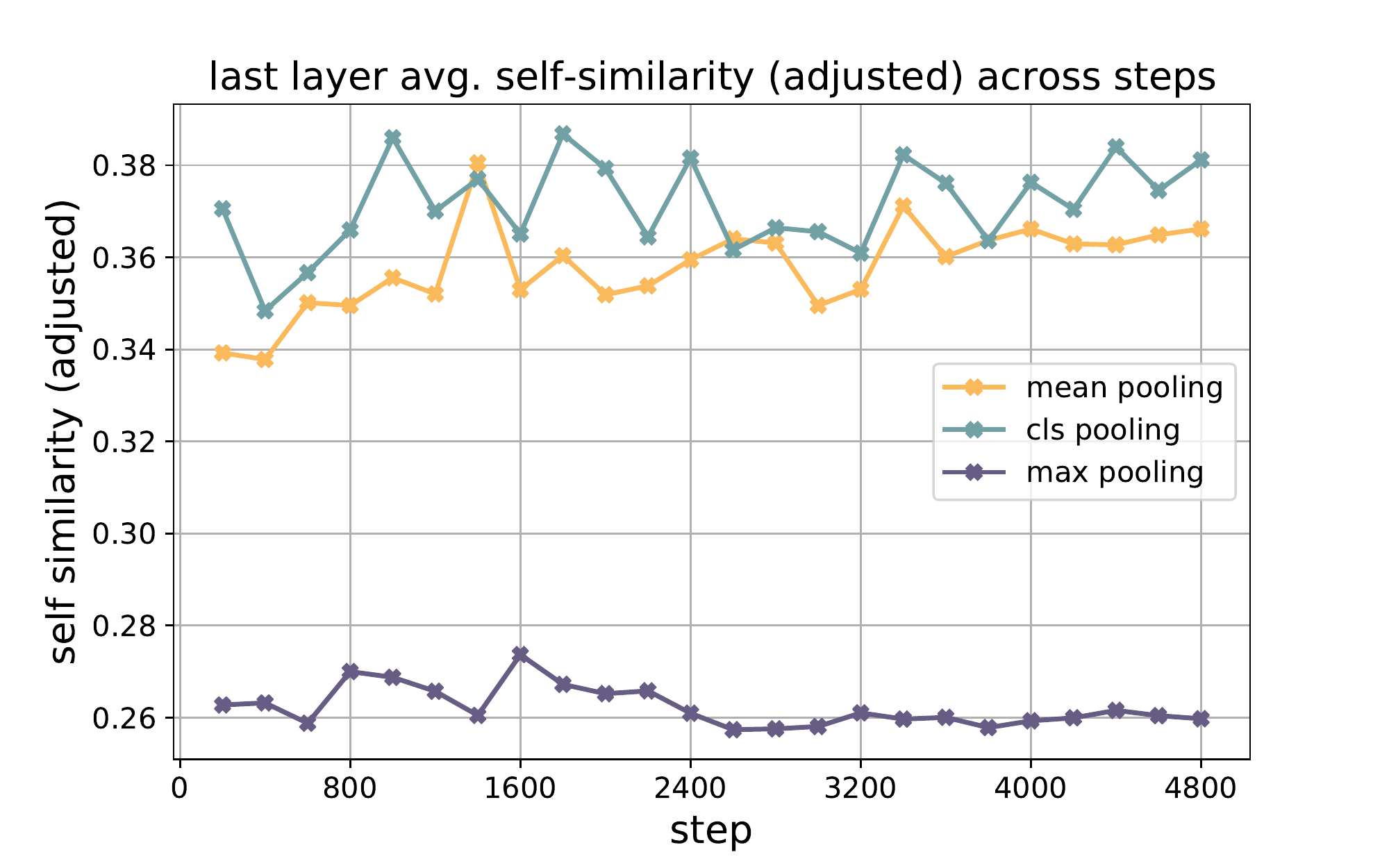}
\caption{Self similarity under different pooling methods, adjusted by anisotropy baseline} \label{temperature-self-adjusted-pooling}
\end{figure}

\begin{figure}[h]
\includegraphics[width=7.7cm]{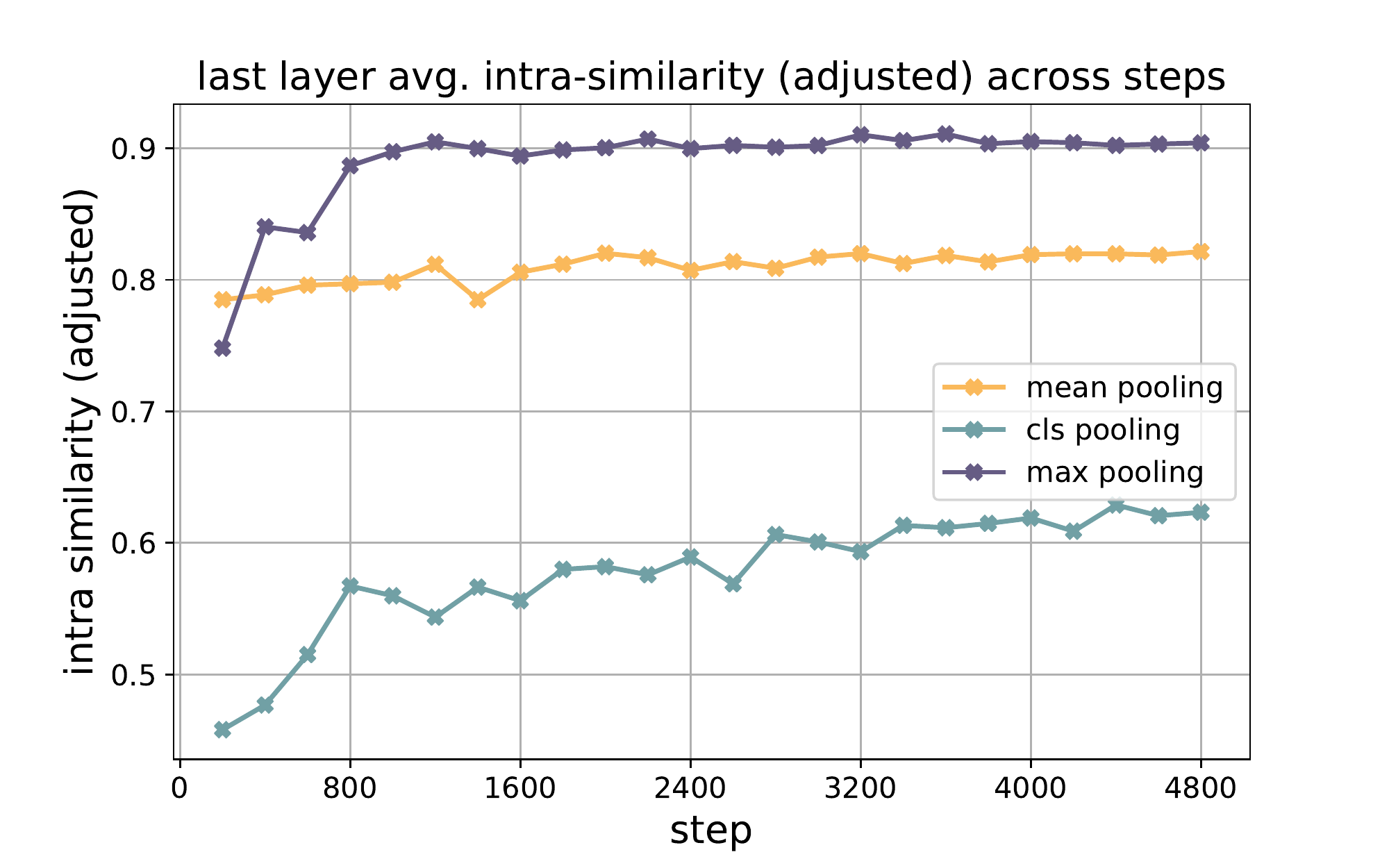}
\caption{Intra-sentence similarity under different pooling methods, adjusted by anisotropy baseline} \label{temperature-intra-adjusted-pooling}
\end{figure}

\section{Self Similarity Change and Correlation across Models}
\label{sec: SSC}
\begin{figure*}[h]
\includegraphics[width=16cm]{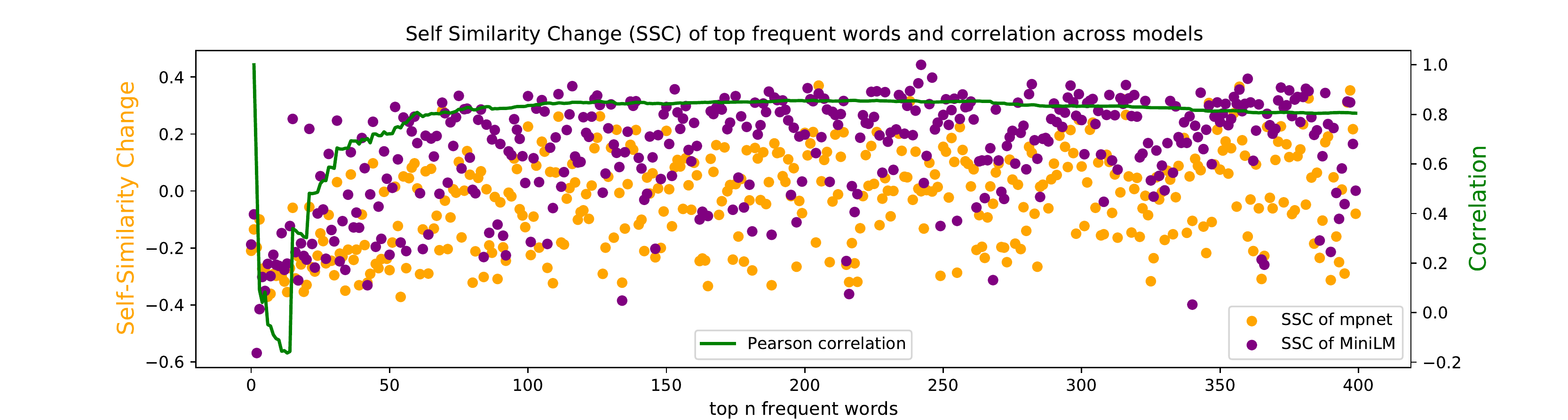}
\caption{Self Similarity Change } 
\label{SSC correlation}
\end{figure*}
In Figure~\ref{SSC correlation} we plot the Self Similarity Change (SSC) across models (mpnet and MiniLM), for the top 400 frequent tokens of the SST-b subset we construct. 

The Pearson correlation between the two accumulated lists of the first $[:n]$ tokens is also plotted. The perfect correlation at the beginning is ignored because the most frequent words at the top are the [pad], [cls] and [sep] tokens. Excluding these, the correlation reaches the peak at 204 as mentioned in the main section, before which the correlation has been slowly stabilized with more tokens considered, while starting to drop after. This shows that the pattern mostly holds for tokens that are above certain frequency, which again provides empirical ground for our analysis on drawing connection of self and intra-sentence similarity to frequency bias.

\end{document}